\documentclass[10pt,twocolumn,letterpaper]{article}
\usepackage{authblk}

\usepackage{cvpr} 
\usepackage{amsmath}
\usepackage{amssymb}
\usepackage{booktabs}
\usepackage{graphicx}
\usepackage{multirow}
\usepackage{tabularx}
\usepackage{array}
\usepackage{multicol} 
\usepackage{placeins} 
\usepackage{xspace}
\usepackage{pgfplots}
\usepgfplotslibrary{groupplots}
\pgfplotsset{compat=1.18}
\definecolor{paperteal}{HTML}{127D80}
\definecolor{papercopper}{HTML}{B46432}

\newcommand{\ctx}{\mathcal{D}_{K}}

\newcommand{\paperTitle}{Scaling to Adapt: Amortized In-Context\\Dynamics Adaptation in Visual World Models}
\newcommand{\MGname}{MechanismGrid\xspace}
\newcommand{\MGabbr}{MG\xspace}
\newcommand{\KDname}{KeyDoor\xspace}
\newcommand{\KDabbr}{KD\xspace}
\newcommand{\CtwoDname}{Continuous2D\xspace}
\newcommand{\CtwoDabbr}{C2D\xspace}
\newcommand{\PointMassname}{PointMass\xspace}
\newcommand{\Reachername}{Reacher\xspace}
\newcommand{\LeWMname}{LeWM\xspace}
\newcommand{\SIGRegname}{SIGReg\xspace}

\newcommand{\KDContextGainMinC}{0}

\newcommand{\KDContextGainMaxC}{75.6591796875}
\newcommand{\MGContextGainMinC}{0}
\newcommand{\MGContextGainPeak}{34.8876953125}
\newcommand{\MGContextGainMaxC}{34.48486328125}
\newcommand{\KDNavigationGainMinC}{-0.78125}

\newcommand{\KDNavigationGainMaxC}{22.65625}

\newcommand{\MGNavigationGainPeak}{45.9375}
\newcommand{\MGNavigationGainMaxC}{26.5625}
\newcommand{\ContextGainPP}[2][1]{%
  \pgfmathprintnumber[fixed,precision=#1]{#2}%
}
\newcommand{\ContextGainDeltaPP}[3][1]{%
  \pgfmathparse{#2-(#3)}%
  \ContextGainPP[#1]{\pgfmathresult}%
}


\ifdefined\mainonly
  \usepackage{xr-hyper}
\fi

\definecolor{cvprblue}{rgb}{0.21,0.49,0.74}
\usepackage[pagebackref,breaklinks,colorlinks,allcolors=cvprblue]{hyperref}
\ifdefined\mainonly
\fi

\def\confName{CVPR}
\def\confYear{2027}

\title{\paperTitle}

\makeatletter
\renewcommand\AB@affilsepx{\quad\protect\Affilfont} 
\makeatother

\author[1]{Yuhao Yan\thanks{Equal Contribution.}\thanks{Work done during internship at Huawei.}}
\author[2]{Honghua Dong\protect\footnotemark[1]}
\author[2]{Guo An}
\author[3]{Zijing Wu\protect\footnotemark[2]}
\author[4,2]{Qi Tian}

\affil[1]{Zhejiang University}
\affil[2]{Huawei}
\affil[3]{University of Science and Technology of China\hspace{10em}}
\affil[4]{Guangdong Laboratory of Artificial Intelligence and Digital Economy (SZ)}

\begin{document}
\maketitle

\begin{abstract}
    Changes in dynamics can alter the consequences of an action without being evident from the current observation. Visual world models therefore need to adapt when deployment dynamics differ from those seen during training. We study this adaptation from an amortized perspective: predictive pretraining learns to use a set of world transitions as context to infer dynamics, and deployment reuses this capability with frozen weights. In two controlled environments, we select model sizes and training-data amounts by training loss at fixed compute budgets, then evaluate context benefits on unseen combinations of familiar mechanisms. We find that the minimum observed training loss is well described by a decreasing power law over the fitted compute range. From the smallest to the largest evaluated compute budget, the accuracy gain from 256 context transitions over no context increases by \ContextGainDeltaPP{\MGContextGainMaxC}{\MGContextGainMinC} percentage points in MechanismGrid and \ContextGainDeltaPP{\KDContextGainMaxC}{\KDContextGainMinC} points in KeyDoor. Additional pretraining can thus strengthen a frozen model's use of fixed context, but lower prediction loss does not guarantee higher navigation success. Our diagnostics suggest that compositional overfitting and limitations in decoding and goal scoring contribute to this gap.
\end{abstract}

\section{Introduction}
\label{sec:introduction}

\begin{figure*}[t]
  \vspace{-0.5em}
  \centering
  \input{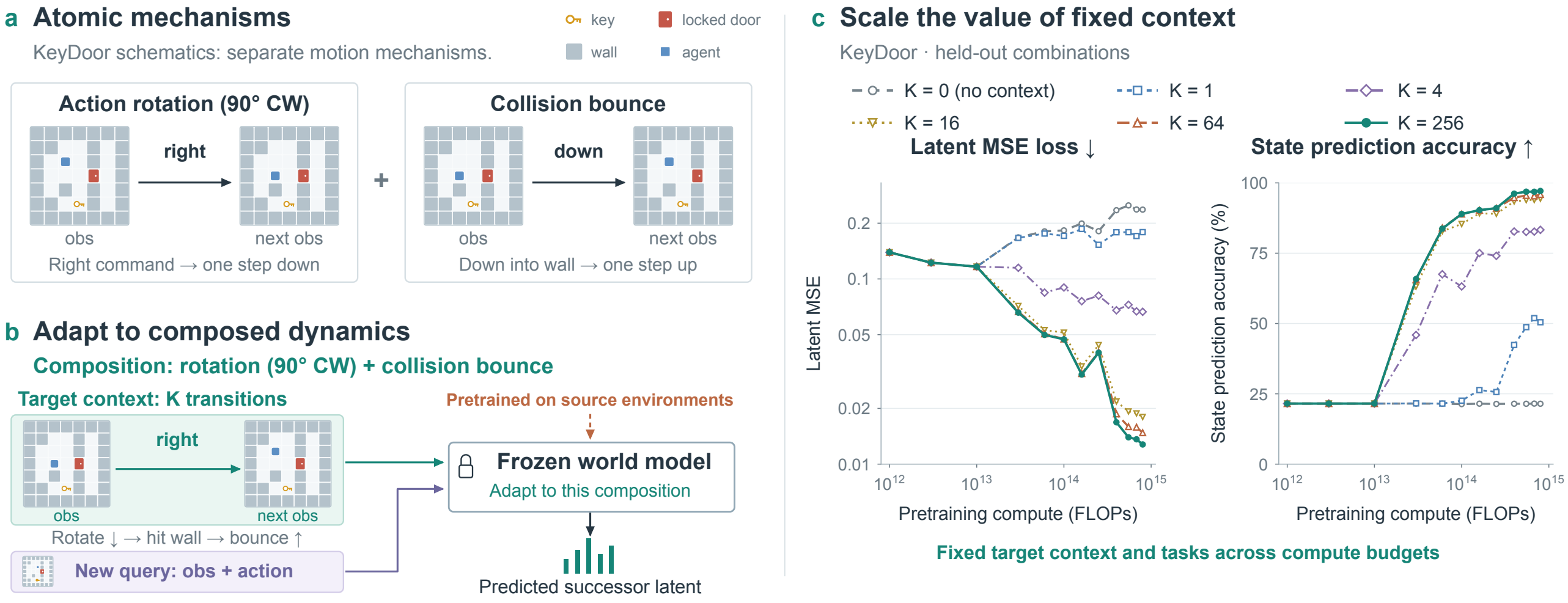}
  \vspace{-1.5em}
  \caption{\textbf{Pretraining shapes how frozen world models use target context.} (a) Atomic mechanisms: KeyDoor (KD) action rotation and collision bounce. (b) Their composition. Context transitions and a separate query share the dynamics. (c) Latent mean squared error (MSE; left) and state prediction accuracy (right) on held-out KD combinations. Each budget uses the model with minimum training loss, fixed across context lengths $K$ in both panels. The benefit over $K=0$ increases beyond the low-compute regime.}
  \vspace{-0.5em}
  \label{fig:overview}
\end{figure*}

Visual world models support planning and control by predicting action consequences \cite{hafner2019planet,hafner2020dreamer}. At deployment, changes in underlying mechanisms can alter those consequences without being apparent from the current observation (Fig.~\ref{fig:overview}a). Observed action outcomes provide the additional evidence needed to adapt predictions to unfamiliar dynamics from a limited set of deployment transitions.

GPT-3 demonstrated \emph{in-context learning} in large language models: a pretrained model can perform tasks specified by prompt examples without parameter updates \cite{brown2020gpt3}. \emph{In-context world models} apply this principle to dynamics prediction, conditioning on observed transitions from the deployment environment \cite{wang2026contextdiversity,luo2026physicalworld}. We refer to the supplied set of transitions as \emph{context}; each record contains an observation, an action, and the next observation. Context-conditioned dynamics learning trains models to use such records to predict separate transitions under the same dynamics \cite{lee2020cadm,evans2022iida}. Training across environments with different dynamics \emph{amortizes} adaptation: pretraining learns a shared computation for using context, reused at deployment with new context and frozen weights (Fig.~\ref{fig:overview}a,b).

This amortized view makes pretraining scale relevant to adaptation. Existing world-model scaling studies relate model size, data, and compute to prediction and downstream performance \cite{pearce2025scaling,zholus2026robojepa}. For context-conditioned models, we ask: \emph{how does more predictive pretraining change the benefit of a fixed context in the target environment?} We measure the accuracy gain from context, since context-free prediction can also improve.

To measure this benefit, we study two controlled visual environments, \MGname (\MGabbr) and \KDname (\KDabbr). Context gains on familiar dynamics may reflect recognition of a training configuration. We therefore hold out combinations of familiar mechanisms, asking whether target experience supports prediction of their joint effects without pretraining on those combinations. At each budget, we select the model--data allocation with lowest training loss, then evaluate with and without context on target transitions, queries, and navigation tasks matched across budgets. Context interventions probe dynamics information; planning comparisons hold each predictor and its search settings fixed to examine decoding and goal scoring.

Over the fitted compute range in each environment, minimum observed training loss is well described by a decreasing power law. From the smallest to the largest evaluated compute budget, the accuracy benefit of 256 context transitions over no context grows by \ContextGainDeltaPP{\MGContextGainMaxC}{\MGContextGainMinC} percentage points in \MGabbr and \ContextGainDeltaPP{\KDContextGainMaxC}{\KDContextGainMinC} in \KDabbr. In \MGabbr, training-loss improvements coexist with high-compute regressions in held-out-combination latent error and navigation, despite a sustained accuracy benefit from context. Lower prediction loss need not translate into higher navigation success rate (SR); realizing predictive gains in control also depends on latent decoding and goal-distance evaluation.

Our contributions are threefold:

\begin{itemize}
  \interlinepenalty=10000
  \item \textbf{Quantifying how pretraining changes context benefits.} We compare model--data allocations at fixed compute and evaluate frozen models with and without matched context. In \MGabbr and \KDabbr, we show stronger prediction benefits from the same context alongside approximate power-law scaling of minimum observed training loss.
  \item \textbf{Demonstrating compositional dynamics adaptation.} We show that \MGabbr models trained on atomic mechanisms predict unseen combinations from RGB transition context with frozen weights. Context interventions show that these gains transfer across layouts with matching dynamics and depend on intact action--outcome pairings.
  \item \textbf{We find that reductions in prediction loss from scaling pretraining do not necessarily translate into higher downstream task success rates.} We use the cross-entropy method (CEM) to select navigation actions by scoring predicted outcomes against the goal. With predictors and search settings fixed, oracle decoding and goal scoring can improve success, yet \MGabbr's high-compute decline persists.
\end{itemize}

\section{Related Work}
\label{sec:related-work}

\paragraph{World models and predictive representations.}
We use \emph{world model} to denote a learned model of environment dynamics that predicts action consequences, here as representations of the next observation in a learned visual latent space. Ha and Schmidhuber's World Models combines visual compression and recurrent dynamics modeling for control \cite{ha2018worldmodels}. Subsequent latent world models support planning and policy learning \cite{hafner2019planet,hafner2020dreamer,hafner2025dreamerv3,hansen2024tdmpc2}. I-JEPA and V-JEPA develop image and video feature-prediction objectives \cite{assran2023ijepa,bardes2024vjepa}; DINO-WM and V-JEPA~2 use pretrained visual representations for dynamics and planning \cite{zhou2025dinowm,assran2025vjepa2}. Our study builds on \LeWMname's joint training of visual representations and latent dynamics with LeJEPA's \SIGRegname regularization \cite{maes2026leworldmodel,balestriero2025lejepa}.

\paragraph{In-context learning and dynamics adaptation.}
GPT-3 demonstrates task adaptation from prompt examples without weight updates \cite{brown2020gpt3}. Distributional, Bayesian, and compression perspectives relate context use to predictive training \cite{chan2022distributional,xie2022bayesian,elmoznino2025occam}. This connects to learning to adapt across tasks \cite{finn2017maml,duan2016rl2,rakelly2019pearl}: Neural Processes learn context-to-query prediction \cite{garnelo2018cnp,kim2019anp}, while in-context RL learns behavior from interaction histories \cite{laskin2022algorithm,lee2023supervisediclrl,grigsby2024amago}. For dynamics, context-conditioned prediction and system identification extract environment information from transitions \cite{lee2020cadm,roder2025dali,zhang2025dynamicsasprompts,piga2024incontextsysid}. IIDA learns this identification through prediction loss without dynamics-parameter labels \cite{evans2022iida}. ICWM uses interaction-derived context to condition robotic policies \cite{wang2026roboticicwm}, while concurrent PhysiCalWorld learns physics descriptors through latent prediction \cite{luo2026physicalworld}. CLAW instead generates low-rank model updates from target transitions \cite{palafox2026claw}. L2World distinguishes environment recognition from environment learning and studies the roles of context length and environment diversity \cite{wang2026contextdiversity}. We study how pretraining compute changes context benefits on held-out mechanism combinations.

\paragraph{Scaling prediction and adaptation.}
Scaling laws relate performance to model size, data, and compute \cite{kaplan2020scaling,hoffmann2022compute}. Sequential ICL studies connect model and data scale to adaptation, while task diversity shapes in-context regression \cite{raparthy2024sequentialicl,raventos2023diversity}. Pearce et al. study pretraining-loss scaling in agents and world models; RoboJEPA relates compute to latent rollout prediction and planning \cite{pearce2025scaling,zholus2026robojepa}. We connect these lines by asking how pretraining scale changes the benefit of fixed context over no context in prediction and control.

\section{Preliminaries}
\label{sec:preliminaries}

\subsection{Agents, Environments, and World Models}
\label{sec:latent-world-models}
An \emph{agent} is a system that chooses actions to accomplish a task. The \emph{environment} is the world or simulator with which it interacts. At step $t$, the environment has a \emph{state} $s_t$, describing its configuration, such as object positions and velocities. The agent receives an \emph{observation} $o_t$, consisting of RGB images in our setting.

An \emph{action} $a_t$ is a command issued by the agent, such as a directional move or a bounded continuous control signal. The environment's \emph{dynamics} specify how the current state and executed action determine the next state or its distribution:
\begin{equation}
    s_{t+1}\sim p_\theta(\cdot\mid s_t,a_t).
    \label{eq:environment-dynamics}
\end{equation}
Each environment $e$ specifies a scene layout and a dynamics setting $\theta$, which remains fixed within a task. After the state changes, the agent receives the next, or \emph{successor}, observation $o_{t+1}$. The tuple $\tau_t=(o_t,a_t,o_{t+1})$ records one observed \emph{transition}; a sequence of such transitions forms a \emph{trajectory}. Figure~\ref{fig:overview}a illustrates how the issued command and resulting motion can differ.

A \emph{world model} learns to predict an action's outcome from an observation and a specified action. We use a \emph{latent world model}, whose encoder $E_\phi$ maps an observation to a learned feature vector $z_t$ and whose predictor $F_\phi$ estimates the next observation's representation:
\begin{equation}
    z_t=E_\phi(o_t),\qquad
    \hat z_{t+1}=F_\phi(z_t,a_t).
    \label{eq:latent-prediction}
\end{equation}
Here $\phi$ contains the encoder and predictor parameters. \LeWMname jointly trains them by matching $\hat z_{t+1}$ to the observed successor embedding $E_\phi(o_{t+1})$. It uses \SIGRegname to prevent representation collapse, in which different observations receive the same uninformative representation \cite{maes2026leworldmodel,balestriero2025lejepa}.

Prediction supplies an outcome for a given action; \emph{control} chooses which action to execute. A navigation task specifies an initial observation $o_0$, a goal observation $g$ depicting the desired outcome, and an action budget limiting the number of actions available to reach the goal. A world model supports control by predicting candidate action outcomes before execution, allowing a planner to evaluate their progress toward the goal. Section~\ref{sec:frozen-planning} describes this planning procedure.

\subsection{In-Context Adaptation and Environment Transfer}
\label{sec:in-context-world-models}
\begin{figure*}[t]
      \vspace{-0.5em}
      \centering
      \input{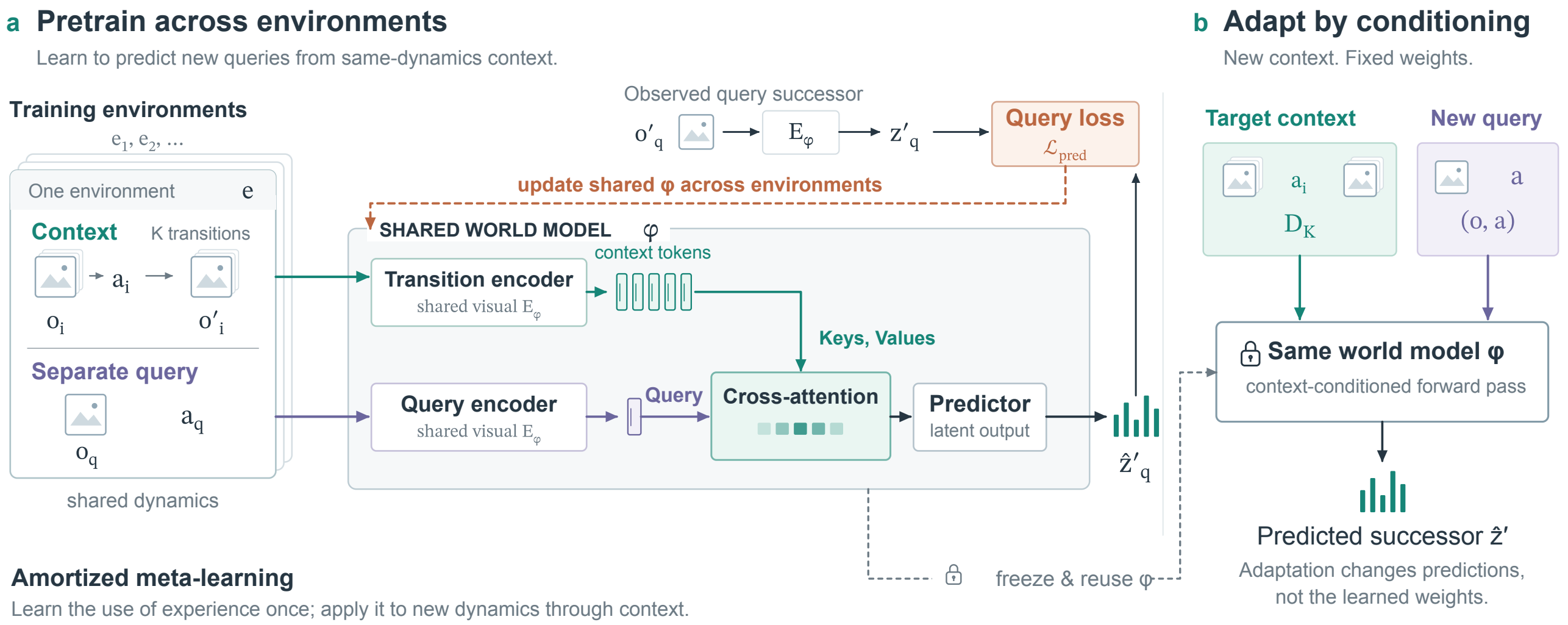}
      \vspace{-2em}
      \caption{\textbf{Learning to use context through prediction.} (a) Random-action transitions form a context set; separate queries share its dynamics. A predictor uses context, the query observation, and its action to predict the successor latent. The observed successor provides the training target through the shared encoder. (b) At deployment, new context changes the frozen model's predictions.}
      \vspace{-0.5em}
      \label{fig:method}
\end{figure*}

An \emph{in-context world model} uses observed transitions as evidence about dynamics not evident from the current observation \cite{evans2022iida,wang2026contextdiversity,luo2026physicalworld}. We call the supplied collection of $K$ transitions \emph{context}. A \emph{query} $(o,a)$ is an observation--action pair whose successor is to be predicted (Fig.~\ref{fig:method}a). In our standard evaluations, context is collected through random actions in the query environment. The context and prediction are
\begin{align}
    \ctx    & =\{(o_i,a_i,o'_i)\}_{i=1}^{K},
    \label{eq:context}                           \\
    \hat z' & =F_\phi\big(E_\phi(o),a,\ctx\big).
    \label{eq:context-pred}
\end{align}
Primes denote the observation or representation after an action. Context successors $o'_i$ are observed inputs; the query successor $o'$ is used only as a training or evaluation target. $K=0$ denotes prediction without transition context.

\emph{Training environments} supply pretraining and model-selection data; \emph{target environments} supply context and evaluation tasks. Pretraining learns shared parameters $\phi$ that define a context-conditioned prediction procedure across training environments. This amortizes \emph{in-context adaptation}: at deployment, the model reuses this learned procedure with target context while keeping $\phi$ fixed (Fig.~\ref{fig:method}b). We use static context, collected before evaluation and held fixed throughout each task.

For navigation, the planner uses these predictions to define a \emph{policy} $\pi$, the agent's rule for selecting actions:
\begin{equation}
    a_t\sim\pi(\cdot\mid o_t,g,\ctx).
    \label{eq:context-policy}
\end{equation}
Thus, context supplies information about the dynamics, while the goal specifies the desired outcome. We evaluate both successor prediction and navigation success to measure predictive adaptation and its use in control.

\section{Learning and Scaling Predictive Adaptation}
\label{sec:method}

We instantiate predictive adaptation by adding transition context to \LeWMname's joint representation and dynamics learning (Fig.~\ref{fig:method}). We then define how to measure its dependence on pretraining scale and connect the frozen predictor to goal-directed control.

\subsection{Amortized Meta-Learning through Prediction}
\label{sec:objective}
In each pretraining episode (Fig.~\ref{fig:method}a), we sample a training environment $e\sim p_{\mathrm{train}}$, a context length $K\sim p_K$, a random-action context set $\ctx$, and a separate query-transition set $\mathcal Q$ under shared dynamics \cite{evans2022iida}.

\paragraph{Context-conditioned visual prediction.}
Using the shared visual encoder $E_\phi$, the transition encoder $T_\phi$ combines observation embeddings and paired spatial features from each record's current and successor images with its action to form a memory token $m_i$. The query token $q$ attends to these tokens, and $P_\phi$ predicts its successor latent:
\begin{align}
      m_i     & =T_\phi(o_i,a_i,o'_i),\qquad
      q=Q_\phi(E_\phi(o),a),\label{eq:transition-query}        \\
      \hat z' & =P_\phi\!\left(q,\operatorname{CrossAttn}_\phi
      \big(q,\{m_i\}_{i=1}^{K}\big)\right).
      \label{eq:attention-pred}
\end{align}
All $K$ complete records enter attention without record positional encodings; observation order within records is preserved. A learned null token supplies memory at $K=0$.

\paragraph{Joint predictive pretraining.}
Following \LeWMname, the observed query successor passes through the same trainable encoder to supply the prediction target:
\begin{equation}
      \mathcal L_{\mathrm{pred}}(\phi;\ctx,\mathcal Q)
      =\frac{1}{|\mathcal Q|}\sum_{(o,a,o')\in\mathcal Q}
      \frac{\|\hat z'-E_\phi(o')\|_2^2}{d_z},
      \label{eq:episode-loss}
\end{equation}
where $d_z$ is the latent dimension. We combine this prediction loss with \SIGRegname on query and observed-successor embeddings to obtain the base objective:
\begin{equation}
      \min_\phi\;\mathbb E_{e,K,\ctx,\mathcal Q}
      \!\left[\mathcal L_{\mathrm{pred}}(\phi;\ctx,\mathcal Q)
            +\lambda\mathcal L_{\mathrm{SIGReg}}\right],
      \label{eq:wm-loss}
\end{equation}
where $\lambda$ is the regularization weight. Gradients through context, query, and target branches jointly learn representations and how to use context across environments. At deployment (Fig.~\ref{fig:method}b), the same encoders turn transitions from the target environment into memory for new queries; all parameters stay frozen. Appendix~\ref{app:training} details architecture and study-specific objectives, including state-attribute supervision from successor RGB and recursive prediction.

\paragraph{Compositional transfer.}
To examine generalization to new combinations of mechanisms, we study \emph{compositional dynamics}, where multiple mechanisms jointly govern action consequences (Fig.~\ref{fig:overview}a,b). We test whether context supports prediction on held-out combinations of familiar mechanisms, without supplying mechanism labels. Training dynamics may be atomic or composed, but test combinations are absent from pretraining. Appendix~\ref{app:environments} specifies the training and test rules.

\subsection{Pretraining Scale and the Value of Context}
\label{sec:scaling-design}
We measure pretraining scale by model size, data, and computation. Model size $N$ counts WM parameters. Data size $D$ counts consumed context--query examples, each pairing a context set with one query transition. Each pair counts once, regardless of context length $K$. Compute $C$ counts profiled forward- and backward-pass floating-point operations (FLOPs), including context processing and auxiliary losses; head training and deployment search are separate costs (Appendix~\ref{app:compute}).

At fixed compute, a larger model costs more per example and can process fewer examples \cite{hoffmann2022compute}. We therefore compare several model sizes at each budget to find the best measured allocation.

\paragraph{Training loss and allocation.}
At budget $C$, let $\mathcal M_C$ be the tested model--data ($N,D$) allocations, all trained to that budget under the same context distribution. A candidate's \emph{terminal training objective}, $\overline{\mathcal L}_{\mathrm{train}}$, is the complete objective---Eq.~\ref{eq:wm-loss} plus any study-specific auxiliary---at the end of training. We select
\begin{equation}
      j^*(C)=\arg\min_{j\in\mathcal M_C}
      \overline{\mathcal L}_{\mathrm{train}}(j),
      \label{eq:allocation-selection}
\end{equation}
$(N^*,D^*,L^*)$ denote its parameter count, number of consumed examples, and terminal training objective.

\paragraph{Power-law fitting.}
We fit each selected quantity independently within each setting:
\begin{equation}
      \widehat{Y^*}(C)=A_Y(C/C_0)^{b_Y},\quad Y\in\{N,D,L\},
      \label{eq:allocation-powers}
\end{equation}
Here $C_0=10^{14}$ FLOPs is a reference scale and $A_Y$ is the fitted value there. Using actual profiled compute, we fit $\log Y^*$ against $\log(C/C_0)$ by least squares, with equal weight per budget. We assess fit accuracy by leaving out one whole budget, refitting on the others, and predicting the omitted value. These are retrospective fits to measured allocations.

\paragraph{Context benefit across compute.}
Let $\phi_C$ be the model selected at budget $C$, frozen for evaluation. We distinguish \emph{train} (consumed training data), \emph{val} (different data under the same mechanisms), and \emph{test} (held-out dynamics). Successor accuracy retrieves the nearest encoded successor-image candidate to each predicted latent and counts a prediction as correct when the retrieved successor matches the ground truth (Appendix~\ref{app:metrics}). Navigation success rate (SR) is the fraction of tasks reaching the goal within an action budget. For metric $m$ on split $s\in\{\mathrm{train},\mathrm{val},\mathrm{test}\}$, let $v_K$ and $v_0$ be its values with $K$ context transitions and no context; accuracy and SR are expressed in percent. We define the \emph{context benefit} as
\begin{equation}
      G_m^s(C,K)=
      \begin{cases}
      100(1-v_K/v_0), & m\in\{\mathrm{MSE},\mathrm{total}\},\\
      v_K-v_0, & m\in\{\mathrm{ACC},\mathrm{SR}\}.
      \end{cases}
      \label{eq:gain}
\end{equation}
Here $\mathrm{MSE}$ denotes next-latent prediction MSE and $\mathrm{total}$ the full objective. Loss benefits are relative percentage reductions ($v_0>0$); accuracy and SR benefits are percentage-point differences. Positive $G$ means context helps; growth with $C$ at fixed $K$ means the same context becomes more beneficial. Section~\ref{sec:experimental-setting} specifies the paired comparisons.

\subsection{From Predictive Adaptation to Control}
\label{sec:frozen-planning}
Inspired by learned goal scoring \cite{li2026reachability}, we train a scalar goal-distance head on training environments after freezing $\phi$:
\begin{equation}
      \hat d=h_\omega\big(\hat z',E_\phi(g)\big).
      \label{eq:distance-head}
\end{equation}
These two latents are its only inputs. We supervise the head with SmoothL1 regression using training-environment distance labels (fewest required actions or geometric separation), which are not needed at deployment (Appendix~\ref{app:control}). Each WM and its own head stay frozen at evaluation.

Starting from $\hat z_t=E_\phi(o_t)$, the planner rolls out candidate sequences $\mathbf a=(a_t,\ldots,a_{t+H-1})$ and scores them by aggregating predicted goal distances:
\begin{align}
      \hat z_{t+h} & =F_\phi(\hat z_{t+h-1},a_{t+h-1},\ctx),
      \label{eq:latent-rollout}                              \\
      J(\mathbf a) & =\operatorname{Agg}_{h=1}^{H}
      h_\omega\big(\hat z_{t+h},E_\phi(g)\big).
      \label{eq:trajectory-cost}
\end{align}
The score uses terminal distance alone or also includes intermediate distances. Following \LeWMname \cite{maes2026leworldmodel}, the cross-entropy method (CEM) refines its action distribution using low-cost sequences \cite{chua2018pets}; imagination queries only the frozen WM. We call CEM planning with the learned scalar goal-distance head Scalar-CEM. The controller then executes one action, encodes the new observation, and replans until reaching the goal or exhausting the action budget.

To diagnose decoding and goal scoring, Oracle-CEM uses the same WM and CEM settings, but decodes each predicted latent to a candidate state using an evaluator-owned image library, then scores that state by its precomputed distance to the goal.

\section{Experiments}
\label{sec:experiments}

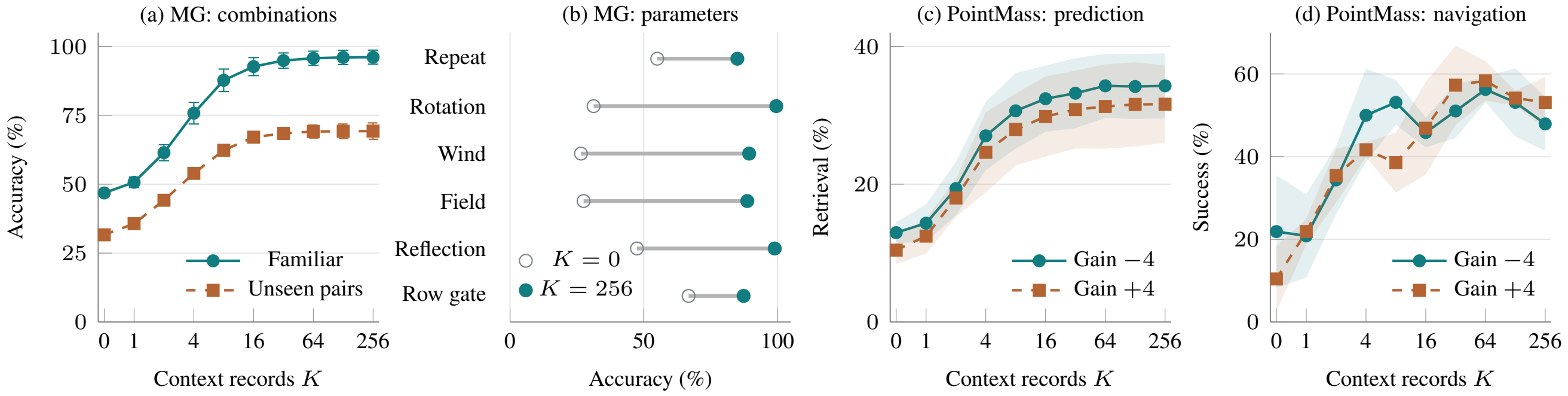
\begin{figure*}[t]
    \vspace{-0.5em}
    \centering
    \definecolor{contextK0}{HTML}{7B858D}
\begin{tikzpicture}
\begin{groupplot}[group style={group size=4 by 1,horizontal sep=1.10cm},
width=0.178\textwidth,height=3.20cm,scale only axis,
tick label style={font=\scriptsize},label style={font=\scriptsize},
every axis title shift=0pt,
title style={font=\scriptsize,at={(.5,1)},anchor=south,yshift=2pt,inner sep=0pt},
axis line style={black!45},tick style={black!45},
axis lines*=left,ymajorgrids=true,grid style={black!10},
legend style={font=\scriptsize,draw=none,fill=none,row sep=-1pt,
at={(0.98,0.04)},anchor=south east,legend columns=1,inner sep=1pt},
xmin=-.2,xmax=9.2,xtick={0,1,3,5,7,9},xticklabels={0,1,4,16,64,256},
xlabel={Context records $K$},ymin=0,ymax=105,ytick={0,25,50,75,100},
ylabel={Accuracy (\%)}]
\nextgroupplot[title={(a) MG: combinations}]
\addplot+[thick,paperteal,mark=*,mark size=1.65pt,mark options={solid,draw=paperteal,fill=paperteal},error bars/.cd,y dir=both,y explicit] coordinates {(0,46.8261719) +- (0,1.05843644) (1,50.7052951) +- (0,1.91403682) (2,61.4355469) +- (0,2.90029695) (3,75.7747396) +- (0,3.9296603) (4,87.6996528) +- (0,4.07483921) (5,92.6812066) +- (0,3.26641724) (6,94.8676215) +- (0,2.77606328) (7,95.7085503) +- (0,2.56823869) (8,96.0026042) +- (0,2.58987859) (9,96.1013455) +- (0,2.54457056)};
\addlegendentry{Familiar}
\addplot+[thick,papercopper,mark=square*,mark size=1.65pt,mark options={solid,draw=papercopper,fill=papercopper},dashed,error bars/.cd,y dir=both,y explicit] coordinates {(0,31.6308594) +- (0,0.711233224) (1,35.6819661) +- (0,1.13282653) (2,44.1927083) +- (0,1.22496651) (3,53.9404297) +- (0,1.5117096) (4,62.3193359) +- (0,1.34973338) (5,67.0605469) +- (0,1.71727204) (6,68.4407552) +- (0,2.12887452) (7,69.0983073) +- (0,2.35391236) (8,69.2154948) +- (0,2.63099029) (9,69.2757161) +- (0,2.97665931)};
\addlegendentry{Unseen pairs}
\nextgroupplot[title={(b) MG: parameters},xshift=0.35cm,
xmin=0,xmax=105,xtick={0,50,100},xticklabels={0,50,100},
xlabel={Accuracy (\%)},ylabel={},ymin=-.55,ymax=5.55,y dir=reverse,
ytick={0,1,2,3,4,5},yticklabels={Repeat,Rotation,Wind,Field,Reflection,Row gate},
ymajorgrids=false,xmajorgrids=true,axis y line=none,axis x line*=bottom,clip=false,
legend style={at={(0.02,0.04)},anchor=south west}]
\addlegendimage{only marks,mark=o,contextK0,mark options={fill=white}}
\addlegendentry{$K=0$}
\addlegendimage{only marks,mark=*,paperteal}
\addlegendentry{$K=256$}
\node[anchor=east,xshift=-4pt,font=\scriptsize] at (axis cs:0,0) {Repeat};
\addplot[black!30,line width=1.2pt,forget plot] coordinates {(55.0238715278,0) (84.9907769097,0)};
\addplot[only marks,mark=o,mark size=2pt,contextK0,mark options={fill=white},forget plot] coordinates {(55.0238715278,0)};
\addplot[only marks,mark=*,mark size=2pt,paperteal,mark options={fill=paperteal},forget plot] coordinates {(84.9907769097,0)};
\node[anchor=east,xshift=-4pt,font=\scriptsize] at (axis cs:0,1) {Rotation};
\addplot[black!30,line width=1.2pt,forget plot] coordinates {(31.2445746528,1) (99.5524088542,1)};
\addplot[only marks,mark=o,mark size=2pt,contextK0,mark options={fill=white},forget plot] coordinates {(31.2445746528,1)};
\addplot[only marks,mark=*,mark size=2pt,paperteal,mark options={fill=paperteal},forget plot] coordinates {(99.5524088542,1)};
\node[anchor=east,xshift=-4pt,font=\scriptsize] at (axis cs:0,2) {Wind};
\addplot[black!30,line width=1.2pt,forget plot] coordinates {(26.523844401,2) (89.4022623698,2)};
\addplot[only marks,mark=o,mark size=2pt,contextK0,mark options={fill=white},forget plot] coordinates {(26.523844401,2)};
\addplot[only marks,mark=*,mark size=2pt,paperteal,mark options={fill=paperteal},forget plot] coordinates {(89.4022623698,2)};
\node[anchor=east,xshift=-4pt,font=\scriptsize] at (axis cs:0,3) {Field};
\addplot[black!30,line width=1.2pt,forget plot] coordinates {(27.4983723958,3) (88.720703125,3)};
\addplot[only marks,mark=o,mark size=2pt,contextK0,mark options={fill=white},forget plot] coordinates {(27.4983723958,3)};
\addplot[only marks,mark=*,mark size=2pt,paperteal,mark options={fill=paperteal},forget plot] coordinates {(88.720703125,3)};
\node[anchor=east,xshift=-4pt,font=\scriptsize] at (axis cs:0,4) {Reflection};
\addplot[black!30,line width=1.2pt,forget plot] coordinates {(47.4609375,4) (98.9664713542,4)};
\addplot[only marks,mark=o,mark size=2pt,contextK0,mark options={fill=white},forget plot] coordinates {(47.4609375,4)};
\addplot[only marks,mark=*,mark size=2pt,paperteal,mark options={fill=paperteal},forget plot] coordinates {(98.9664713542,4)};
\node[anchor=east,xshift=-4pt,font=\scriptsize] at (axis cs:0,5) {Row gate};
\addplot[black!30,line width=1.2pt,forget plot] coordinates {(66.7643229167,5) (87.3291015625,5)};
\addplot[only marks,mark=o,mark size=2pt,contextK0,mark options={fill=white},forget plot] coordinates {(66.7643229167,5)};
\addplot[only marks,mark=*,mark size=2pt,paperteal,mark options={fill=paperteal},forget plot] coordinates {(87.3291015625,5)};
\nextgroupplot[title={(c) PointMass: prediction},ymax=42,ytick={0,20,40,60},ylabel={Retrieval (\%)}]
\addplot[draw=none,fill=paperteal,fill opacity=.12,forget plot] coordinates {(0,14.4961415832) (1,17.0623549398) (2,23.2521932335) (3,31.9634681282) (4,36.1474751471) (5,37.2146941852) (6,38.2749077733) (7,38.9334581599) (8,38.8963428421) (9,39.0046023022) (9,29.5500851978) (8,29.4630321579) (7,29.6212293401) (6,28.1313422267) (5,27.5965037314) (4,25.1480977696) (3,22.0729902051) (2,15.4847859331) (1,11.6160304769) (0,11.4478688335)} \closedcycle;
\addplot+[thick,paperteal,mark=*,mark size=1.65pt,mark options={solid,draw=paperteal,fill=paperteal}] coordinates {(0,12.9720052) (1,14.3391927) (2,19.3684896) (3,27.0182292) (4,30.6477865) (5,32.405599) (6,33.203125) (7,34.2773438) (8,34.1796875) (9,34.2773438)};
\addlegendentry{Gain $-4$}
\addplot[draw=none,fill=papercopper,fill opacity=.12,forget plot] coordinates {(0,12.4652331695) (1,15.0104199169) (2,20.7358224013) (3,30.4403680108) (4,33.1669970483) (5,35.6395963376) (6,36.4804226056) (7,37.4019325064) (8,37.6976279349) (9,37.2082849326) (9,26.0078609007) (8,25.4534137317) (7,25.1631716603) (6,25.1732232277) (5,23.9958203291) (4,22.724930035) (3,18.8109340725) (2,15.2667817654) (1,9.92447591646) (0,8.40065224715)} \closedcycle;
\addplot+[thick,papercopper,mark=square*,mark size=1.65pt,mark options={solid,draw=papercopper,fill=papercopper},dashed] coordinates {(0,10.4329427) (1,12.4674479) (2,18.0013021) (3,24.625651) (4,27.9459635) (5,29.8177083) (6,30.8268229) (7,31.2825521) (8,31.5755208) (9,31.6080729)};
\addlegendentry{Gain $+4$}
\nextgroupplot[title={(d) PointMass: navigation},ymax=70,ytick={0,20,40,60},ylabel={Success (\%)}]
\addplot[draw=none,fill=paperteal,fill opacity=.12,forget plot] coordinates {(0,35.4965591986) (1,30.878802876) (2,42.6429728471) (3,61.2673477358) (4,58.5376587737) (5,49.4417725158) (6,57.546872915) (7,59.375) (8,61.3929728471) (9,54.421872915) (9,41.4114604183) (8,44.8570271529) (7,53.125) (6,44.5364604183) (5,42.2248941509) (4,47.7123412263) (3,38.7326522642) (2,26.1070271529) (1,10.7878637906) (0,8.25344080144)} \closedcycle;
\addplot+[thick,paperteal,mark=*,mark size=1.65pt,mark options={solid,draw=paperteal,fill=paperteal}] coordinates {(0,21.875) (1,20.8333333) (2,34.375) (3,50) (4,53.125) (5,45.8333333) (6,51.0416667) (7,56.25) (8,53.125) (9,47.9166667)};
\addlegendentry{Gain $-4$}
\addplot[draw=none,fill=papercopper,fill opacity=.12,forget plot] coordinates {(0,18.2810775367) (1,25) (2,41.921872915) (3,43.4708862579) (4,45.7585450315) (5,58.1423477358) (6,66.8386993645) (7,63.1068496822) (8,55.9708862579) (9,59.375) (9,46.875) (8,52.3624470754) (7,53.5598169844) (6,47.7446339688) (5,35.6076522642) (4,31.3247883018) (3,39.8624470754) (2,28.9114604183) (1,18.75) (0,2.55225579659)} \closedcycle;
\addplot+[thick,papercopper,mark=square*,mark size=1.65pt,mark options={solid,draw=papercopper,fill=papercopper},dashed] coordinates {(0,10.4166667) (1,21.875) (2,35.4166667) (3,41.6666667) (4,38.5416667) (5,46.875) (6,57.2916667) (7,58.3333333) (8,54.1666667) (9,53.125)};
\addlegendentry{Gain $+4$}
\end{groupplot}
\end{tikzpicture}
    \vspace{-1.5em}
    \caption{\textbf{Context improves prediction and navigation with frozen model weights.}
        \textbf{(a)} MechanismGrid (MG) successor-state accuracy across $K$ context transitions after training on atomic mechanisms; curves compare familiar rules and unseen mechanism pairs.
        \textbf{(b)} A separate MG parameter-transfer study; rows average targets within each mechanism family.
        \textbf{(c,d)} PointMass with held-out force gains: (c) successor retrieval among 64 candidate images sharing the observed frame and differing in the future frame; (d) goal-reaching success with learned Scalar-CEM.}
    \vspace{-0.5em}
    \label{fig:icl-prediction}
\end{figure*}

\subsection{Experimental Setting}
\label{sec:experimental-setting}

\MGname (\MGabbr) and \KDname (\KDabbr) test adaptation to hidden action rules from RGB. \MGabbr tests atomic-to-pair transfer (training on individual mechanisms and testing unseen pairs) and context interventions. \KDabbr tests unseen combinations after training on atomic and composed rules, on maps where one key unlocks the door. We vary pretraining compute in both environments. In separate studies on \MGabbr, \CtwoDname (\CtwoDabbr), and \PointMassname, we test adaptation to dynamics parameter settings absent from training (Appendix~\ref{app:environments}).

\MGabbr and \CtwoDabbr use Eq.~\ref{eq:wm-loss} with $\lambda=0.09$; \KDabbr adds auxiliary supervision for position and keys derived from successor RGB, and \PointMassname uses two-step recursive prediction. Appendix~\ref{app:training} details objectives and training context lengths; Appendix~\ref{app:context} describes context collection. Table~\ref{tab:protocol} records model selection and training-seed coverage.

For scaling, each budget uses the model with the lowest terminal training objective, averaged over the final window of up to 100 training steps. Context transitions, prediction queries, and navigation tasks are matched across compute budgets; queries and tasks are also fixed across $K$, with shorter contexts forming prefixes of longer ones. Each WM and its head remain frozen across $K$, with fixed planner settings and SIGReg projections for loss comparisons (Appendix~\ref{app:metrics}).

We report total loss (the full objective), next-latent prediction MSE, successor accuracy, and navigation SR on the train, val, and test splits where measured. Test dynamics comprise held-out mechanism combinations or parameter values, as specified above. Discrete accuracy requires a correct complete successor state; continuous accuracy uses 64-image retrieval. \PointMassname uses consecutive-frame pairs; successor candidates share the query's current frame and vary only the future frame. Accuracy complements MSE because encoder scales differ across models. Error bars denote training-seed standard deviations unless specified otherwise (Appendix~\ref{app:metrics}).

\subsection{In-Context Generalization}
\label{sec:icl-results}

\subsubsection{Prediction under Unseen Dynamics}
\label{sec:transfer-results}

\textbf{RQ1: Can a frozen model use context to generalize to unseen dynamics?}
Context improves prediction of unseen mechanism combinations without weight updates. After training only on individual \MGabbr mechanisms, unseen-pair accuracy rises from 31.6\% to 69.3\% as $K$ increases from 0 to 256, with the same queries at every $K$ (Fig.~\ref{fig:icl-prediction}a).

\begin{figure*}[t]
    \vspace{-0.5em}
    \centering
    \input{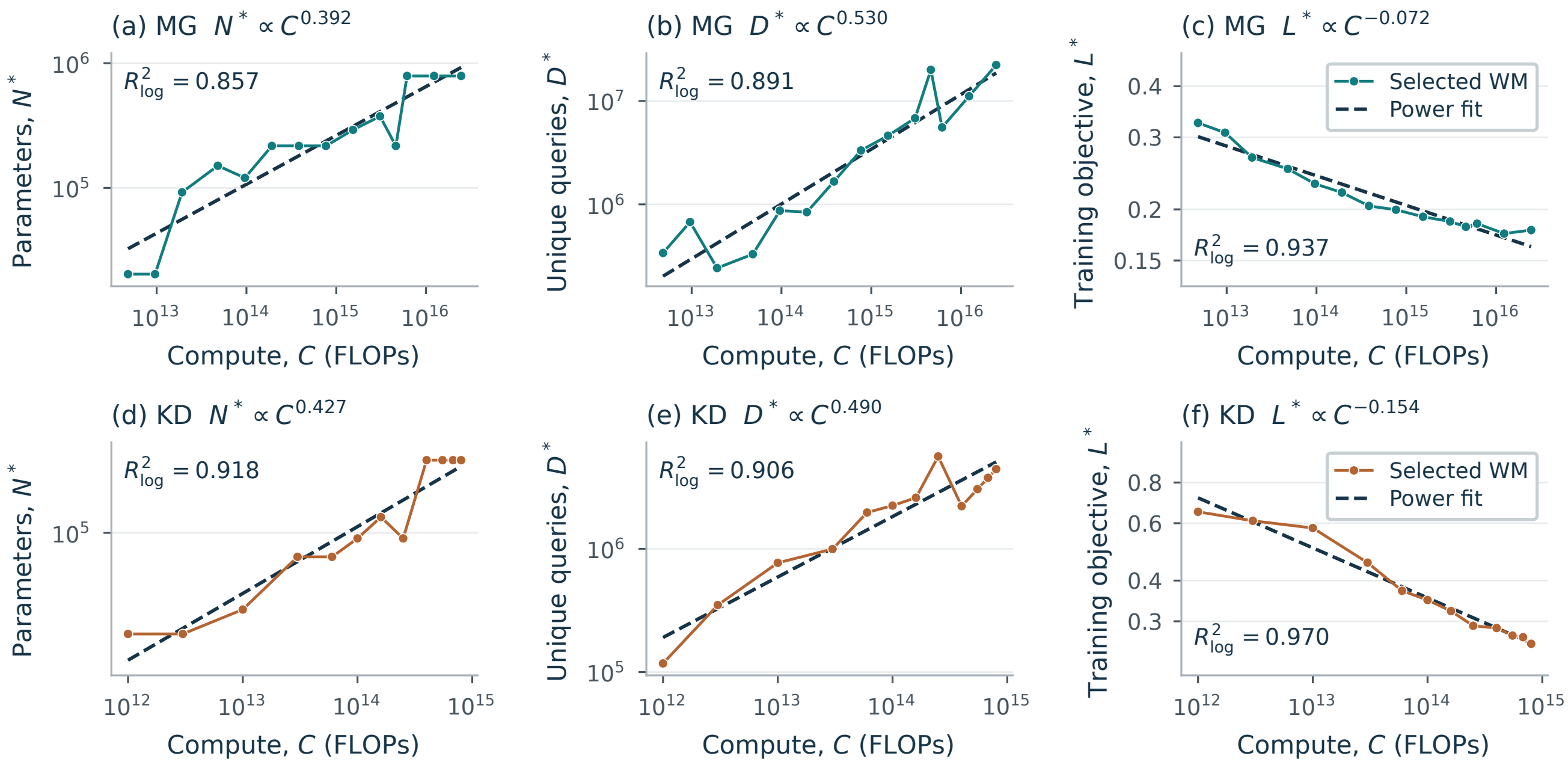}
    \vspace{-1.5em}
    \caption{\textbf{Training loss and model--data allocation at fixed compute. Power-law fits describe the measured budgets.} In MechanismGrid (MG) and KeyDoor (KD), $N^*,D^*,L^*$ are the parameter count, number of consumed context--query examples, and terminal objective of the model with lowest training loss at each compute budget $C$. $L$ combines latent prediction MSE and representation regularization, plus KD's successor-state auxiliary derived from RGB.}
    \vspace{-0.5em}
    \label{fig:scaling-ndl}
    \label{fig:mg-scaling-fit}
    \label{fig:kd-scaling-fit}
\end{figure*}

\begin{figure*}[t]
    \vspace{-0.5em}
    \centering
    \input{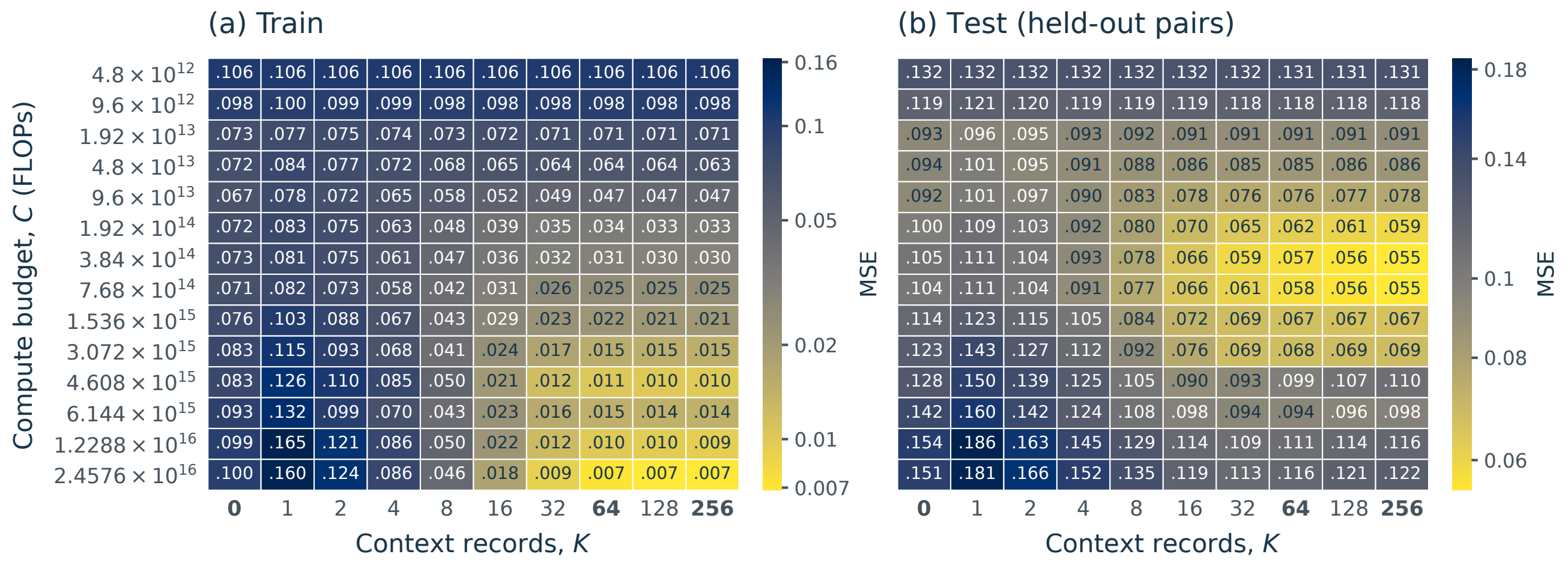}
    \vspace{-2em}
    \caption{\textbf{MG context use emerges and improves before high-compute regressions.} Cells show MechanismGrid (MG) next-latent mean squared error (MSE) on consumed training queries (Train) and held-out mechanism pairs (Test). Each compute budget fixes the model with lowest terminal training loss across panels and context lengths $K$. At large $K$, training MSE falls overall while test MSE first falls and then rises with compute. Bold $K$ values were sampled in training. Panels use separate logarithmic color scales.}
    \vspace{-0.5em}
    \label{fig:mg-train-dev-mse}
\end{figure*}

The benefit extends to unseen \MGabbr parameter values (Fig.~\ref{fig:icl-prediction}b). In this parameter-transfer study, the WM outperforms a matched transition-retrieval baseline using the same encoder and context, tuned on val data (Appendix~\ref{app:extended}).

Context improves successor-image retrieval on all six \CtwoDabbr parameter targets. In \PointMassname, accuracy averaged over the two unseen force gains rises from 11.7\% at $K=0$ to 32.9\% at $K=256$ (Table~\ref{tab:continuous}; Fig.~\ref{fig:icl-prediction}c). Goal-reaching success also improves between these endpoints, although it fluctuates at intermediate $K$ (Fig.~\ref{fig:icl-prediction}d). These results extend the context benefit to continuous actions and force changes.

For the unseen \MGabbr row-gate rule, overall accuracy improves, but accuracy falls on queries where the gate should block horizontal movement: the model often predicts motion through the gate (Appendix~\ref{app:extended}).

\subsubsection{Effects of Context}
\label{sec:context-results}

\textbf{RQ2: Which information in context enables adaptation?}
We next isolate the context information behind these gains: matching dynamics and intact action--outcome correspondences. In controlled \MGabbr interventions, context from the correct rule remains useful across layouts, whereas context from a wrong rule on the query layout performs below the no-context baseline (Table~\ref{tab:interventions}). This points to action consequences rather than scene similarity alone as the useful evidence.

\begin{table}[t]
    \centering\small
    \vspace{-0.5em}
    \caption{\textbf{Context benefit depends on dynamics and action--outcome correspondence.} MechanismGrid (MG) successor-state accuracy and learned Scalar-CEM success on unseen mechanism pairs. Across interventions, models, queries, and navigation tasks stay fixed; nonempty contexts contain $K=256$ transitions. This cohort is separate from Fig.~\ref{fig:icl-prediction}a.}
    \vspace{-0.5em}
    \label{tab:interventions}
    \begin{tabular}{@{}lrr@{}}
\toprule
Context & Accuracy $\uparrow$ & Success $\uparrow$ \\
 & (\%) & (\%) \\
\midrule
No context & 31.8 & 7.7 \\
Matching rule, same layout & 69.1 & 42.3 \\
Matching rule, different layout & 69.1 & 40.8 \\
Wrong rule, same layout & 22.0 & 2.4 \\
Shift action assignments & 40.2 & 7.4 \\
Shift successor assignments & 29.2 & 6.5 \\
Repeat one record & 36.8 & 8.2 \\
Reverse record order & 69.1 & 42.3 \\
\bottomrule
\end{tabular}

    \vspace{-0.5em}
\end{table}

Reassigning actions or successor images across context records reduces the benefit while preserving the underlying images and actions. Repeating a single valid transition provides little benefit: more records need not provide more information. Reversing the order of complete transition records preserves performance, consistent with context attention without record positional encodings. Planning seeds are also fixed across interventions (Appendix~\ref{app:context}).

\subsection{Effects of Pretraining Scale}
\label{sec:scaling-results}

\subsubsection{Training Loss and Model--Data Allocation}
\label{sec:allocation-results}

\textbf{RQ3: How do pretraining resources shape loss and model--data allocation?}
We turn from context content to pretraining loss and allocation. Power laws fit the minimum observed training loss $L^*$ over the measured budgets in both \MGabbr and \KDabbr (Fig.~\ref{fig:scaling-ndl}). The fits are less accurate for selected model size $N^*$ and query count $D^*$ (Table~\ref{tab:scaling-fit-errors}). Over the selected model-size ranges, measured pretraining compute is approximated by $C\propto N^{1.2}D$: the cost per context--query example is about $9\times10^7(N/10^5)^{1.2}$ FLOPs in \MGabbr and $5\times10^7(N/10^5)^{1.2}$ in \KDabbr. Appendix~\ref{app:compute-cost} details this finite-range cost approximation and the fit diagnostics.

\subsubsection{Context Benefit across Pretraining Budgets}
\label{sec:compute-prediction}

\textbf{RQ4: Does more pretraining increase the prediction benefit of a fixed target context?}
We now evaluate how the models selected by training loss use fixed test context. $G_{\mathrm{ACC}}^{\mathrm{test}}(C,256)$ generally grows with compute, increasing by \ContextGainDeltaPP{\MGContextGainMaxC}{\MGContextGainMinC} percentage points in \MGabbr and \ContextGainDeltaPP{\KDContextGainMaxC}{\KDContextGainMinC} in \KDabbr from the smallest to the largest evaluated budget (Eq.~\ref{eq:gain}; Appendix~\ref{app:prediction-confirmation}). In \MGabbr, the largest-compute gain of \ContextGainPP{\MGContextGainMaxC} percentage points remains close to its peak of \ContextGainPP{\MGContextGainPeak}.

\textbf{KD: context becomes useful as compute increases.} $G_{\mathrm{MSE}}^{\mathrm{test}}(C,256)$ is near zero at low compute and generally increases at larger budgets (Fig.~\ref{fig:overview}c, left). The curves for different context lengths separate as compute increases; beyond this transition, conditioned MSE broadly improves while the no-context predictor remains weak. Across the smallest tested budgets, adding records changes prediction little.

\begingroup\clubpenalty=10000
\textbf{MG: accuracy gains alongside rising test MSE.} Figure~\ref{fig:mg-prediction-accuracy} shows accuracy rising from 30.5\% to 69.1\% at $K=256$, compared with 30.5\% to 34.6\% at $K=0$. The $K=64/256$ curves improve overall with local regressions and flatten near 70\% at high compute. Despite this trend, Fig.~\ref{fig:mg-train-dev-mse} shows high-compute regressions in test MSE. As compute increases, $G_{\mathrm{MSE}}^{\mathrm{train}}(C,256)$ grows overall, while $G_{\mathrm{MSE}}^{\mathrm{test}}(C,256)$ peaks at intermediate compute and declines overall at higher budgets. At $K=256$, five of six fixed-capacity settings show rising late-training test MSE while train and val MSE fall (Fig.~\ref{fig:mg-mse-training}). Together, these observations support compositional overfitting: further training can increase latent prediction error on held-out combinations, despite a positive context benefit.
\par\endgroup

\subsection{Navigation Success across Pretraining Budgets}
\label{sec:scoring-results}

\textbf{RQ5: Do predictive gains from more pretraining translate into successful control?}
Given the differing accuracy and latent-MSE trends in RQ4, we test whether the accuracy gains support navigation by comparing Scalar-CEM with Oracle-CEM using the same WMs and search settings (Sec.~\ref{sec:frozen-planning}). In \KDabbr, learned SR is near zero at the smallest budgets and generally rises with fluctuations at larger budgets. Its context benefit $G_{\mathrm{SR}}^{\mathrm{test}}(C,256)$ increases by \ContextGainDeltaPP{\KDNavigationGainMaxC}{\KDNavigationGainMinC} percentage points between the compute endpoints. Oracle-CEM follows the same overall upward trend and reaches nearly $100\%$ at $K=256$ at larger budgets (Fig.~\ref{fig:navigation-compute-control}b). This gap points to the prediction-to-goal-scoring interface as a control bottleneck, although the oracle changes both decoding and distances.

\begin{figure}[t]
    \centering
\begingroup
\definecolor{accgray}{HTML}{77848C}
\definecolor{accblue}{HTML}{4C82B8}
\definecolor{accpurple}{HTML}{8870AA}
\definecolor{accgold}{HTML}{B09532}
\begin{tikzpicture}
\begin{axis}[
  width=0.97\linewidth,height=3.65cm,
  xmode=log,xmin=3e12,xmax=3.4e16,xtick={1e13,1e14,1e15,1e16},
  ymin=25,ymax=75,ytick={30,40,50,60,70},
  xlabel={Pretraining compute (FLOPs)},ylabel={Acc. (\%)},
  tick label style={font=\scriptsize,black!65},
  label style={font=\footnotesize},
  axis line style={black!45},tick style={black!45},
  axis x line*=bottom,axis y line*=left,
  ymajorgrids,grid style={black!12},
  legend style={font=\scriptsize,draw=none,fill=none,
    at={(0.5,1.035)},anchor=south,legend columns=3,
    /tikz/every even column/.append style={column sep=6pt}},
  legend cell align=left,
  clip=false,
]
\node[anchor=north west,font=\scriptsize\bfseries] at (rel axis cs:0.02,0.96) {MG};
\addplot+[color=accgray,mark=triangle,mark size=1.8pt,mark options={fill=white,draw=accgray},dashed,line width=0.95pt]
 table[x=actual_C,y=K0,col sep=comma] {figures/mg_prediction_accuracy_compute.csv};
\addlegendentry{$K=0$}
\addplot+[color=accblue,mark=square,mark size=1.6pt,mark options={fill=white,draw=accblue},densely dashed,line width=0.95pt]
 table[x=actual_C,y=K1,col sep=comma] {figures/mg_prediction_accuracy_compute.csv};
\addlegendentry{$K=1$}
\addplot+[color=accpurple,mark=diamond,mark size=1.8pt,mark options={fill=white,draw=accpurple},dashdotted,line width=0.95pt]
 table[x=actual_C,y=K4,col sep=comma] {figures/mg_prediction_accuracy_compute.csv};
\addlegendentry{$K=4$}
\addplot+[color=accgold,mark=triangle,mark size=1.8pt,mark options={rotate=180,fill=white,draw=accgold},dotted,line width=0.95pt]
 table[x=actual_C,y=K16,col sep=comma] {figures/mg_prediction_accuracy_compute.csv};
\addlegendentry{$K=16$}
\addplot+[color=papercopper,mark=triangle,mark size=1.8pt,mark options={fill=white,draw=papercopper},dashed,line width=0.95pt]
 table[x=actual_C,y=K64,col sep=comma] {figures/mg_prediction_accuracy_compute.csv};
\addlegendentry{$K=64$}
\addplot+[color=paperteal,mark=*,mark size=1.6pt,mark options={fill=paperteal,draw=paperteal},solid,line width=1.2pt]
 table[x=actual_C,y=K256,col sep=comma] {figures/mg_prediction_accuracy_compute.csv};
\addlegendentry{$K=256$}
\end{axis}
\end{tikzpicture}
\endgroup
    \vspace{-1em}
    \caption{\textbf{MG prediction accuracy improves with compute and sufficient context.} Complete successor-state accuracy on held-out mechanism pairs, using the minimum-training-loss WM at each of 14 budgets. Each model uses the same 8,192 queries across $K$.}
    \label{fig:mg-prediction-accuracy}
    \vspace{0.5em}
\begin{tikzpicture}
\begin{groupplot}[
  group style={group size=1 by 2,vertical sep=0.68cm},
  width=0.97\linewidth,height=3.65cm,
  xmode=log,ymin=0,ymax=104,ytick={0,25,50,75,100},
  ylabel={SR (\%)},
  tick label style={font=\scriptsize,black!65},
  label style={font=\footnotesize},
  axis line style={black!45},tick style={black!45},
  axis x line*=bottom,axis y line*=left,
  ymajorgrids,grid style={black!12},
  legend style={font=\scriptsize,draw=none,fill=none,
    at={(0.28,1.035)},anchor=south,legend columns=2,
    /tikz/every even column/.append style={column sep=4pt}},
  legend cell align=left,
  clip=false,
]
\nextgroupplot[xmin=3e12,xmax=3.4e16,xtick={1e13,1e14,1e15,1e16}]
\node[anchor=north west,font=\scriptsize\bfseries] at (rel axis cs:0.02,0.96) {(a) MG};
\addplot+[color=paperteal,mark=triangle,mark size=1.8pt,mark options={fill=white,draw=paperteal},dashed,line width=0.95pt]
 table[x=actual_C,y=scalar_K0,col sep=comma] {figures/navigation_mg_compute.csv};
\addlegendentry{Scalar, $K=0$}
\addplot+[color=paperteal,mark=*,mark size=1.6pt,mark options={fill=paperteal,draw=paperteal},line width=1.2pt]
 table[x=actual_C,y=scalar_K256,col sep=comma] {figures/navigation_mg_compute.csv};
\addlegendentry{Scalar, $K=256$}
\addplot+[color=papercopper,mark=diamond,mark size=1.8pt,mark options={fill=white,draw=papercopper},dashed,line width=0.95pt]
 table[x=actual_C,y=oracle_K0,col sep=comma] {figures/navigation_mg_compute.csv};
\addlegendentry{Oracle, $K=0$}
\addplot+[color=papercopper,mark=square*,mark size=1.5pt,mark options={fill=papercopper,draw=papercopper},line width=1.2pt]
 table[x=actual_C,y=oracle_K256,col sep=comma] {figures/navigation_mg_compute.csv};
\addlegendentry{Oracle, $K=256$}
\nextgroupplot[xmin=7e11,xmax=1.2e15,xtick={1e12,1e13,1e14,1e15},
 xlabel={Pretraining compute (FLOPs)}]
\node[anchor=north west,font=\scriptsize\bfseries] at (rel axis cs:0.02,0.96) {(b) KD};
\addplot+[color=paperteal,mark=triangle,mark size=1.8pt,mark options={fill=white,draw=paperteal},dashed,line width=0.95pt,forget plot]
 table[x=actual_C,y=scalar_K0,col sep=comma] {figures/navigation_kd_compute.csv};
\addplot+[color=paperteal,mark=*,mark size=1.6pt,mark options={fill=paperteal,draw=paperteal},line width=1.2pt,forget plot]
 table[x=actual_C,y=scalar_K256,col sep=comma] {figures/navigation_kd_compute.csv};
\addplot+[color=papercopper,mark=diamond,mark size=1.8pt,mark options={fill=white,draw=papercopper},dashed,line width=0.95pt,forget plot]
 table[x=actual_C,y=oracle_K0,col sep=comma] {figures/navigation_kd_compute.csv};
\addplot+[color=papercopper,mark=square*,mark size=1.5pt,mark options={fill=papercopper,draw=papercopper},line width=1.2pt,forget plot]
 table[x=actual_C,y=oracle_K256,col sep=comma] {figures/navigation_kd_compute.csv};
\end{groupplot}
\end{tikzpicture}
    \vspace{-0.5em}
    \caption{\textbf{Context gains in navigation success rate (SR) need not grow monotonically with compute.} (a) MechanismGrid (MG); (b) KeyDoor (KD). At each budget, the WM with lowest training loss is shared across both planners and context lengths $K$. Oracle-CEM uses the same search settings as learned Scalar-CEM, with privileged latent decoding and goal distances. At high compute and $K=256$, MG's decline persists under Oracle-CEM, whereas KD's oracle SR stays near 100\%.}
    \label{fig:navigation-compute-control}
    \vspace{-0.5em}
\end{figure}

In \MGabbr, learned Scalar-CEM context benefit $G_{\mathrm{SR}}^{\mathrm{test}}(C,256)$ rises to a peak of \ContextGainPP{\MGNavigationGainPeak} percentage points and then trends downward to \ContextGainPP{\MGNavigationGainMaxC} at the largest compute budget. The same overall rise and post-peak decline persist under Oracle-CEM, which uses the same WMs with privileged decoding and distances (Fig.~\ref{fig:navigation-compute-control}a). The regression under both planners coincides with rising test MSE at higher compute (Fig.~\ref{fig:mg-train-dev-mse}). The oracle advantage quantifies how much performance can improve with privileged decoding and goal distances (Appendix~\ref{app:control}).

\section{Discussion and Conclusion}
\label{sec:limitations}
\label{sec:conclusion}

In this work, we studied how predictive pretraining shapes in-context adaptation in visual world models. Across two controlled environments, minimum observed training loss follows an approximate power law over the fitted compute range, while more pretraining can increase the prediction and control benefits of fixed context on unseen mechanism combinations. These findings support an amortized view of adaptation, in which pretraining develops the ability to use new experience with frozen weights.

Compositional overfitting and the oracle advantage reveal two limits of our \LeWMname-style framework: fitting training dynamics does not ensure transfer, and useful predictions do not ensure effective action selection. Our compute-scaling evidence is also confined to controlled discrete environments.

Future work should examine which training distributions and prediction objectives sustain adaptation as compute grows, and test whether these gains extend to visually richer continuous environments with partial observability and longer-horizon control.

\section*{LLM Usage Statement}

OpenAI's ChatGPT and Codex (using GPT models, including \texttt{gpt-6.1-sol} and \texttt{gpt-6-astra}) were used in the following two ways.

\paragraph{Paper writing.}
Across all sections, manuscript writing primarily involved interactive generation with LLMs, followed by human review and approval.

\paragraph{Code development and experiments.}
The authors designed the code and provided instructions and experimental protocols. LLMs modified the code and executed experiments, with the authors reviewing and approving the changes and verifying the experimental outputs.

The research ideas, methodological decisions, experimental interpretations, and conclusions were developed and determined by the authors. Experimental configurations, data processing procedures, and reported results were systematically verified against experimental records to ensure data integrity, result accuracy, and reproducibility.

{
  \footnotesize
  \bibliographystyle{ieeenat_fullname}
  \bibliography{main}

\begin{thebibliography}{40}
\providecommand{\natexlab}[1]{#1}
\providecommand{\url}[1]{\texttt{#1}}
\expandafter\ifx\csname urlstyle\endcsname\relax
  \providecommand{\doi}[1]{doi: #1}\else
  \providecommand{\doi}{doi: \begingroup \urlstyle{rm}\Url}\fi

\bibitem[Assran et~al.(2023)Assran, Duval, Misra, Bojanowski, Vincent, Rabbat,
  LeCun, and Ballas]{assran2023ijepa}
Mahmoud Assran, Quentin Duval, Ishan Misra, Piotr Bojanowski, Pascal Vincent,
  Michael Rabbat, Yann LeCun, and Nicolas Ballas.
\newblock {Self-Supervised Learning from Images with a Joint-Embedding
  Predictive Architecture}.
\newblock In \emph{IEEE/CVF Conference on Computer Vision and Pattern
  Recognition}, pages 15619--15629, 2023.

\bibitem[Assran et~al.(2025)Assran, Bardes, Fan, Garrido, Howes, Komeili,
  Muckley, Rizvi, Roberts, Sinha, Zholus, Arnaud, Gejji, Martin, Hogan, Dugas,
  Bojanowski, Khalidov, Labatut, Massa, Szafraniec, Krishnakumar, Li, Ma,
  Chandar, Meier, LeCun, Rabbat, and Ballas]{assran2025vjepa2}
Mido Assran, Adrien Bardes, David Fan, Quentin Garrido, Russell Howes, Mojtaba
  Komeili, Matthew Muckley, Ammar Rizvi, Claire Roberts, Koustuv Sinha, Artem
  Zholus, Sergio Arnaud, Abha Gejji, Ada Martin, Francois~Robert Hogan, Daniel
  Dugas, Piotr Bojanowski, Vasil Khalidov, Patrick Labatut, Francisco Massa,
  Marc Szafraniec, Kapil Krishnakumar, Yong Li, Xiaodong Ma, Sarath Chandar,
  Franziska Meier, Yann LeCun, Michael Rabbat, and Nicolas Ballas.
\newblock {V-JEPA 2: Self-Supervised Video Models Enable Understanding,
  Prediction and Planning}.
\newblock arXiv preprint arXiv:2506.09985, 2025.

\bibitem[Balestriero and LeCun(2025)]{balestriero2025lejepa}
Randall Balestriero and Yann LeCun.
\newblock {LeJEPA: Provable and Scalable Self-Supervised Learning Without the
  Heuristics}.
\newblock arXiv preprint arXiv:2511.08544, 2025.

\bibitem[Bardes et~al.(2024)Bardes, Garrido, Ponce, Chen, Rabbat, LeCun,
  Assran, and Ballas]{bardes2024vjepa}
Adrien Bardes, Quentin Garrido, Jean Ponce, Xinlei Chen, Michael Rabbat, Yann
  LeCun, Mido Assran, and Nicolas Ballas.
\newblock {Revisiting Feature Prediction for Learning Visual Representations
  from Video}.
\newblock \emph{Transactions on Machine Learning Research}, 2024.

\bibitem[Brown et~al.(2020)Brown, Mann, Ryder, Subbiah, Kaplan, Dhariwal,
  Neelakantan, Shyam, Sastry, Askell, Agarwal, Herbert-Voss, Krueger, Henighan,
  Child, Ramesh, Ziegler, Wu, Winter, Hesse, Chen, Sigler, Litwin, Gray, Chess,
  Clark, Berner, McCandlish, Radford, Sutskever, and Amodei]{brown2020gpt3}
Tom~B. Brown, Benjamin Mann, Nick Ryder, Melanie Subbiah, Jared Kaplan,
  Prafulla Dhariwal, Arvind Neelakantan, Pranav Shyam, Girish Sastry, Amanda
  Askell, Sandhini Agarwal, Ariel Herbert-Voss, Gretchen Krueger, Tom Henighan,
  Rewon Child, Aditya Ramesh, Daniel~M. Ziegler, Jeffrey Wu, Clemens Winter,
  Christopher Hesse, Mark Chen, Eric Sigler, Mateusz Litwin, Scott Gray,
  Benjamin Chess, Jack Clark, Christopher Berner, Sam McCandlish, Alec Radford,
  Ilya Sutskever, and Dario Amodei.
\newblock {Language Models are Few-Shot Learners}.
\newblock In \emph{Advances in Neural Information Processing Systems}, pages
  1877--1901, 2020.

\bibitem[Chan et~al.(2022)Chan, Santoro, Lampinen, Wang, Singh, Richemond,
  McClelland, and Hill]{chan2022distributional}
Stephanie C.~Y. Chan, Adam Santoro, Andrew~K. Lampinen, Jane~X. Wang,
  Aaditya~K. Singh, Pierre~H. Richemond, James~L. McClelland, and Felix Hill.
\newblock {Data Distributional Properties Drive Emergent In-Context Learning in
  Transformers}.
\newblock In \emph{Advances in Neural Information Processing Systems}, pages
  18878--18891, 2022.

\bibitem[Chua et~al.(2018)Chua, Calandra, McAllister, and Levine]{chua2018pets}
Kurtland Chua, Roberto Calandra, Rowan McAllister, and Sergey Levine.
\newblock {Deep Reinforcement Learning in a Handful of Trials using
  Probabilistic Dynamics Models}.
\newblock In \emph{Advances in Neural Information Processing Systems}, 2018.

\bibitem[Duan et~al.(2016)Duan, Schulman, Chen, Bartlett, Sutskever, and
  Abbeel]{duan2016rl2}
Yan Duan, John Schulman, Xi Chen, Peter~L. Bartlett, Ilya Sutskever, and Pieter
  Abbeel.
\newblock {RL$^2$: Fast Reinforcement Learning via Slow Reinforcement
  Learning}.
\newblock arXiv preprint arXiv:1611.02779, 2016.

\bibitem[Elmoznino et~al.(2025)Elmoznino, Marty, Kasetty, Gagnon, Mittal,
  Fathi, Sridhar, and Lajoie]{elmoznino2025occam}
Eric Elmoznino, Tom Marty, Tejas Kasetty, Leo Gagnon, Sarthak Mittal, Mahan
  Fathi, Dhanya Sridhar, and Guillaume Lajoie.
\newblock {In-Context Learning and Occam's Razor}.
\newblock In \emph{Proceedings of the 42nd International Conference on Machine
  Learning}, pages 15296--15319. PMLR, 2025.

\bibitem[Evans et~al.(2022)Evans, Thankaraj, and Pinto]{evans2022iida}
Ben Evans, Abitha Thankaraj, and Lerrel Pinto.
\newblock {Context is Everything: Implicit Identification for Dynamics
  Adaptation}.
\newblock In \emph{IEEE International Conference on Robotics and Automation},
  pages 2642--2648. IEEE, 2022.

\bibitem[Finn et~al.(2017)Finn, Abbeel, and Levine]{finn2017maml}
Chelsea Finn, Pieter Abbeel, and Sergey Levine.
\newblock {Model-Agnostic Meta-Learning for Fast Adaptation of Deep Networks}.
\newblock In \emph{International Conference on Machine Learning}, pages
  1126--1135, 2017.

\bibitem[Garnelo et~al.(2018)Garnelo, Rosenbaum, Maddison, Ramalho, Saxton,
  Shanahan, Teh, Rezende, and Eslami]{garnelo2018cnp}
Marta Garnelo, Dan Rosenbaum, Christopher Maddison, Tiago Ramalho, David
  Saxton, Murray Shanahan, Yee~Whye Teh, Danilo Rezende, and S.~M.~Ali Eslami.
\newblock {Conditional Neural Processes}.
\newblock In \emph{International Conference on Machine Learning}, pages
  1704--1713, 2018.

\bibitem[Grigsby et~al.(2024)Grigsby, Fan, and Zhu]{grigsby2024amago}
Jake Grigsby, Linxi Fan, and Yuke Zhu.
\newblock {AMAGO: Scalable In-Context Reinforcement Learning for Adaptive
  Agents}.
\newblock In \emph{International Conference on Learning Representations}, 2024.

\bibitem[Ha and Schmidhuber(2018)]{ha2018worldmodels}
David Ha and J{\"u}rgen Schmidhuber.
\newblock {World Models}.
\newblock arXiv preprint arXiv:1803.10122, 2018.

\bibitem[Hafner et~al.(2019)Hafner, Lillicrap, Fischer, Villegas, Ha, Lee, and
  Davidson]{hafner2019planet}
Danijar Hafner, Timothy Lillicrap, Ian Fischer, Ruben Villegas, David Ha,
  Honglak Lee, and James Davidson.
\newblock {Learning Latent Dynamics for Planning from Pixels}.
\newblock In \emph{International Conference on Machine Learning}, pages
  2555--2565, 2019.

\bibitem[Hafner et~al.(2020)Hafner, Lillicrap, Ba, and
  Norouzi]{hafner2020dreamer}
Danijar Hafner, Timothy Lillicrap, Jimmy Ba, and Mohammad Norouzi.
\newblock {Dream to Control: Learning Behaviors by Latent Imagination}.
\newblock In \emph{International Conference on Learning Representations}, 2020.

\bibitem[Hafner et~al.(2025)Hafner, Pasukonis, Ba, and
  Lillicrap]{hafner2025dreamerv3}
Danijar Hafner, Jurgis Pasukonis, Jimmy Ba, and Timothy Lillicrap.
\newblock {Mastering diverse control tasks through world models}.
\newblock \emph{Nature}, 640:\penalty0 647--653, 2025.

\bibitem[Hansen et~al.(2024)Hansen, Su, and Wang]{hansen2024tdmpc2}
Nicklas Hansen, Hao Su, and Xiaolong Wang.
\newblock {TD-MPC2: Scalable, Robust World Models for Continuous Control}.
\newblock In \emph{International Conference on Learning Representations}, 2024.

\bibitem[Hoffmann et~al.(2022)Hoffmann, Borgeaud, Mensch, Buchatskaya, Cai,
  Rutherford, de~Las~Casas, Hendricks, Welbl, Clark, Hennigan, Noland,
  Millican, van~den Driessche, Damoc, Guy, Osindero, Simonyan, Elsen, Vinyals,
  Rae, and Sifre]{hoffmann2022compute}
Jordan Hoffmann, Sebastian Borgeaud, Arthur Mensch, Elena Buchatskaya, Trevor
  Cai, Eliza Rutherford, Diego de Las~Casas, Lisa~Anne Hendricks, Johannes
  Welbl, Aidan Clark, Thomas Hennigan, Eric Noland, Katherine Millican, George
  van~den Driessche, Bogdan Damoc, Aurelia Guy, Simon Osindero, Kar{\'e}n
  Simonyan, Erich Elsen, Oriol Vinyals, Jack Rae, and Laurent Sifre.
\newblock {An empirical analysis of compute-optimal large language model
  training}.
\newblock In \emph{Advances in Neural Information Processing Systems}, pages
  30016--30030. Curran Associates, Inc., 2022.

\bibitem[Kaplan et~al.(2020)Kaplan, McCandlish, Henighan, Brown, Chess, Child,
  Gray, Radford, Wu, and Amodei]{kaplan2020scaling}
Jared Kaplan, Sam McCandlish, Tom Henighan, Tom~B. Brown, Benjamin Chess, Rewon
  Child, Scott Gray, Alec Radford, Jeffrey Wu, and Dario Amodei.
\newblock {Scaling Laws for Neural Language Models}.
\newblock arXiv preprint arXiv:2001.08361, 2020.

\bibitem[Kim et~al.(2019)Kim, Mnih, Schwarz, Garnelo, Eslami, Rosenbaum,
  Vinyals, and Teh]{kim2019anp}
Hyunjik Kim, Andriy Mnih, Jonathan Schwarz, Marta Garnelo, Ali Eslami, Dan
  Rosenbaum, Oriol Vinyals, and Yee~Whye Teh.
\newblock {Attentive Neural Processes}.
\newblock In \emph{International Conference on Learning Representations}, 2019.

\bibitem[Laskin et~al.(2023)Laskin, Wang, Oh, Parisotto, Spencer, Steigerwald,
  Strouse, Hansen, Filos, Brooks, Gazeau, Sahni, Singh, and
  Mnih]{laskin2022algorithm}
Michael Laskin, Luyu Wang, Junhyuk Oh, Emilio Parisotto, Stephen Spencer,
  Richie Steigerwald, DJ Strouse, Steven Hansen, Angelos Filos, Ethan Brooks,
  Maxime Gazeau, Himanshu Sahni, Satinder Singh, and Volodymyr Mnih.
\newblock {In-context Reinforcement Learning with Algorithm Distillation}.
\newblock In \emph{International Conference on Learning Representations}, 2023.

\bibitem[Lee et~al.(2023)Lee, Xie, Pacchiano, Chandak, Finn, Nachum, and
  Brunskill]{lee2023supervisediclrl}
Jonathan Lee, Annie Xie, Aldo Pacchiano, Yash Chandak, Chelsea Finn, Ofir
  Nachum, and Emma Brunskill.
\newblock {Supervised Pretraining Can Learn In-Context Reinforcement Learning}.
\newblock In \emph{Advances in Neural Information Processing Systems}, pages
  43057--43083, 2023.

\bibitem[Lee et~al.(2020)Lee, Seo, Lee, Lee, and Shin]{lee2020cadm}
Kimin Lee, Younggyo Seo, Seunghyun Lee, Honglak Lee, and Jinwoo Shin.
\newblock {Context-aware Dynamics Model for Generalization in Model-Based
  Reinforcement Learning}.
\newblock In \emph{International Conference on Machine Learning}, pages
  5757--5766, 2020.

\bibitem[Li et~al.(2026)Li, Wang, Qiu, Xiong, and Liu]{li2026reachability}
Liangyu Li, Shengzhi Wang, Libin Qiu, Mingliang Xiong, and Qingwen Liu.
\newblock {World Model Control by Trajectory Reachability Metrics}.
\newblock arXiv preprint arXiv:2605.22164, 2026.

\bibitem[Luo et~al.(2026)Luo, He, Dai, Nam, Fu, Chai, Fazeli, and
  Park]{luo2026physicalworld}
Yunhao Luo, Wentao He, Yinpei Dai, Hyelin Nam, Hongze Fu, Joyce Chai, Nima
  Fazeli, and Jeong~Joon Park.
\newblock {PhysiCalWorld: In-Context Physics Calibration for World Models}.
\newblock arXiv preprint arXiv:2610.30125, 2026.

\bibitem[Maes et~al.(2026)Maes, Le~Lidec, Scieur, LeCun, and
  Balestriero]{maes2026leworldmodel}
Lucas Maes, Quentin Le~Lidec, Damien Scieur, Yann LeCun, and Randall
  Balestriero.
\newblock {LeWorldModel: Stable End-to-End Joint-Embedding Predictive
  Architecture from Pixels}.
\newblock arXiv preprint arXiv:2603.19312, 2026.

\bibitem[Palafox and Fridovich-Keil(2026)]{palafox2026claw}
Fernando Palafox and David Fridovich-Keil.
\newblock {Amortized Low-Rank Adaptation for Model-Based Reinforcement
  Learning}.
\newblock arXiv preprint arXiv:2609.12278, 2026.

\bibitem[Pearce et~al.(2025)Pearce, Rashid, Bignell, Georgescu, Devlin, and
  Hofmann]{pearce2025scaling}
Tim Pearce, Tabish Rashid, David Bignell, Raluca Georgescu, Sam Devlin, and
  Katja Hofmann.
\newblock {Scaling Laws for Pre-training Agents and World Models}.
\newblock In \emph{International Conference on Machine Learning}, pages
  48542--48562, 2025.

\bibitem[Piga et~al.(2024)Piga, Pura, and Forgione]{piga2024incontextsysid}
Dario Piga, Filippo Pura, and Marco Forgione.
\newblock {On the adaptation of in-context learners for system identification}.
\newblock \emph{IFAC-PapersOnLine}, 58\penalty0 (15):\penalty0 277--282, 2024.

\bibitem[Rakelly et~al.(2019)Rakelly, Zhou, Finn, Levine, and
  Quillen]{rakelly2019pearl}
Kate Rakelly, Aurick Zhou, Chelsea Finn, Sergey Levine, and Deirdre Quillen.
\newblock {Efficient Off-Policy Meta-Reinforcement Learning via Probabilistic
  Context Variables}.
\newblock In \emph{International Conference on Machine Learning}, pages
  5331--5340, 2019.

\bibitem[Raparthy et~al.(2024)Raparthy, Hambro, Kirk, Henaff, and
  Raileanu]{raparthy2024sequentialicl}
Sharath~Chandra Raparthy, Eric Hambro, Robert Kirk, Mikael Henaff, and Roberta
  Raileanu.
\newblock {Generalization to New Sequential Decision Making Tasks with
  In-Context Learning}.
\newblock In \emph{International Conference on Machine Learning}, pages
  42138--42158. PMLR, 2024.

\bibitem[Ravent{\'o}s et~al.(2023)Ravent{\'o}s, Paul, Chen, and
  Ganguli]{raventos2023diversity}
Allan Ravent{\'o}s, Mansheej Paul, Feng Chen, and Surya Ganguli.
\newblock {Pretraining task diversity and the emergence of non-Bayesian
  in-context learning for regression}.
\newblock In \emph{Advances in Neural Information Processing Systems}, pages
  14228--14246, 2023.

\bibitem[R{\"o}der et~al.(2025)R{\"o}der, Benad, Eppe, and
  Banerjee]{roder2025dali}
Frank R{\"o}der, Jan Benad, Manfred Eppe, and Pradeep~Kr. Banerjee.
\newblock {Dynamics-Aligned Latent Imagination in Contextual World Models for
  Zero-Shot Generalization}.
\newblock In \emph{Advances in Neural Information Processing Systems}, 2025.

\bibitem[Wang et~al.(2026{\natexlab{a}})Wang, Chen, Zhong, Zheng, Shao, Yu,
  Liu, Wang, Ding, Cao, and Kang]{wang2026contextdiversity}
Fan Wang, Zhiyuan Chen, Yuxuan Zhong, Sunjian Zheng, Pengtao Shao, Bo Yu,
  Shaoshan Liu, Jianan Wang, Ning Ding, Yang Cao, and Yu Kang.
\newblock {Context and Diversity Matter: The Emergence of In-Context Learning
  in World Models}.
\newblock In \emph{International Conference on Learning Representations},
  2026{\natexlab{a}}.

\bibitem[Wang et~al.(2026{\natexlab{b}})Wang, Shi, Fei, Fu, Ji, Gong, and
  Qiu]{wang2026roboticicwm}
Siyin Wang, Junhao Shi, Senyu Fei, Zhaoyang Fu, Li Ji, Jingjing Gong, and
  Xipeng Qiu.
\newblock {In-Context World Modeling for Robotic Control}.
\newblock arXiv preprint arXiv:2606.26025, 2026{\natexlab{b}}.

\bibitem[Xie et~al.(2022)Xie, Raghunathan, Liang, and Ma]{xie2022bayesian}
Sang~Michael Xie, Aditi Raghunathan, Percy Liang, and Tengyu Ma.
\newblock {An Explanation of In-context Learning as Implicit Bayesian
  Inference}.
\newblock In \emph{International Conference on Learning Representations}, 2022.

\bibitem[Zhang et~al.(2025)Zhang, Liu, Huang, Han, Lyu, Xu, and
  Zhao]{zhang2025dynamicsasprompts}
Xilun Zhang, Shiqi Liu, Peide Huang, William~Jongwon Han, Yiqi Lyu, Mengdi Xu,
  and Ding Zhao.
\newblock {Dynamics as Prompts: In-Context Learning for Sim-to-Real System
  Identifications}.
\newblock \emph{IEEE Robotics and Automation Letters}, 10\penalty0
  (4):\penalty0 3190--3197, 2025.

\bibitem[Zholus et~al.(2026)Zholus, Beltran-Velez, Yuan, Chandar, Nagarajan,
  Severo, Sinha, Drozdzal, Soriano, Bohg, Ballas, and
  Assran]{zholus2026robojepa}
Artem Zholus, Nicolas Beltran-Velez, Jianhao Yuan, Sarath Chandar, Tushar
  Nagarajan, Daniel Severo, Koustuv Sinha, Michal Drozdzal, Adriana~Romero
  Soriano, Jeannette Bohg, Nicolas Ballas, and Mahmoud Assran.
\newblock {RoboJEPA: Scaling Robotic Latent World Models}.
\newblock arXiv preprint arXiv:2610.10515, 2026.

\bibitem[Zhou et~al.(2025)Zhou, Pan, LeCun, and Pinto]{zhou2025dinowm}
Gaoyue Zhou, Hengkai Pan, Yann LeCun, and Lerrel Pinto.
\newblock {DINO-WM: World Models on Pre-trained Visual Features enable
  Zero-shot Planning}.
\newblock In \emph{Proceedings of the 42nd International Conference on Machine
  Learning}, pages 79115--79135. PMLR, 2025.

\end{thebibliography}
}

\ifdefined\mainonly\else
  \clearpage
  \appendix
\counterwithin{figure}{section}
\counterwithin{table}{section}
\renewcommand{\theHfigure}{appendix.\thefigure}
\renewcommand{\theHtable}{appendix.\thetable}
\raggedbottom
\makeatletter
\newenvironment{AppendixProtocols}{%
  \renewcommand{\paragraph}{\@startsection{paragraph}{4}{\z@}%
    {4pt plus 1pt minus 1pt}{-1em}{\normalfont\normalsize\bfseries}}%
}{\par}
\makeatother

\newsavebox{\TrainingTestSettings}
\begin{AppendixProtocols}
  \section{Environments and Evaluation Protocols}
  \label{app:environments}

  \subsection{Dynamics and Test Tasks}

  \subsubsection{Discrete environments}
  Agents use five commands (up, down, left, right, and wait) to reach a goal. The model receives actions and $64\times64$ RGB images.

  \paragraph{MechanismGrid (MG).}
  Motion rules change how a command moves the agent on a $17\times17$ obstacle grid (Fig.~\ref{fig:mg-rgb-transitions}). In Fig.~\ref{fig:icl-prediction}(a), models train on single rules and are tested on new maps with familiar rules or unseen pairs. In panel~(b), each model group is tested on one rule parameter withheld from training, such as a rotation angle. Table~\ref{tab:training-test-settings} lists the splits. Figure~\ref{fig:mg-higher-order-details} covers three- to five-rule combinations.

  \paragraph{KeyDoor (KD).}
  On a $15\times15$ grid with sealed walls, collecting a key unlocks its matching door and adds an inventory badge to the RGB image (Fig.~\ref{fig:kd-continuous-rgb-transitions}, top rows). The main experiments use one key and hold out combinations of rotation, reflection, and bounce.

  \subsubsection{Continuous environments}
  \paragraph{Continuous2D (C2D).}
  A point navigates an obstacle map with two action components in $[-1,1]$. Tests use unseen horizontal scales, drifts, and rotations that apply on the right half of the map, as well as five pairs and three triples. Each combination has its own trained models. Figures~\ref{fig:kd-continuous-rgb-transitions} and~\ref{fig:c2d-remaining-rgb-transitions} illustrate all these test rules.

  \paragraph{PointMass and Reacher.}
  PointMass steers a mass, and Reacher guides the tip of a two-link arm. Both test actuator gains absent from training (Table~\ref{tab:training-test-settings}). Each RGB observation stacks two successive frames. One action lasts 25 simulator steps in PointMass and 12 in Reacher.

  \subsection{Data Splits and Training Seeds}
  \label{app:protocols}
  For ICL, val uses rules seen in training. Train/val/test use 40/5/64 layouts per rule for MG parameter transfer and 64/16/32 for other grid and C2D studies. PointMass and Reacher use 32/8/32 and 128/32/32 initializations per rule, respectively. Each test layout or initialization supplies 64 prediction queries. Table~\ref{tab:protocol} counts independently trained WMs.

  Scaling uses the MG single-rule training and unseen-pair test split, and the one-key KD combination split in Table~\ref{tab:training-test-settings}. The train banks contain 5,000 MG and 2,200 KD layouts. Each environment has 100 layouts in val and 100 in test. Each prediction measurement uses 8,192 queries per split and context size (Appendix~\ref{app:strict-prediction}). The scaling test rules are unseen in train. These test results were inspected in earlier analyses.

  \noindent\begin{minipage}{\columnwidth}
    \captionsetup{hypcap=false}\captionof{table}{\textbf{Independently trained models by study.} Each WM seed sets initialization and training randomness. One group can be tested on several rules. The last row counts the selected scaling models in Table~\ref{tab:strict-models}.}
    \label{tab:protocol}\label{tab:comparison-overview}
    \footnotesize\setlength{\tabcolsep}{3pt}
    \renewcommand{\tabularxcolumn}[1]{m{#1}}
    \begin{tabularx}{\columnwidth}{@{}l>{\raggedright\arraybackslash}X@{}}
      \toprule
      Study                       & Model/seed coverage                                         \\
      \midrule
      MG familiar / pairs         & Prediction: 5 / 3 seeds; matched navigation: 3.             \\
      \addlinespace[3pt]
      MG parameter / higher-order & 13 / 9 independent groups, 3 WMs each.               \\
      \addlinespace[3pt]
      KD                          & 3 WMs per ICL setting.                                      \\
      \addlinespace[3pt]
      C2D                         & Three parameter families, 8 combination groups; 3 WMs each. \\
      \addlinespace[3pt]
      PointMass / Reacher         & 3 WMs per simulator, shared across both gains.                 \\
      \addlinespace[3pt]
      MG / KD scaling             & 14 / 12 budgets, one WM seed per budget.                    \\
      \bottomrule
    \end{tabularx}
  \end{minipage}\par\smallskip

\end{AppendixProtocols}
  \begin{lrbox}{\TrainingTestSettings}
    \begin{minipage}{\textwidth}
      \captionsetup{hypcap=false}\captionof{table}{\textbf{Rules used for training and testing.} $R_\theta$ rotates a grid command clockwise by $\theta$ degrees. A plus sign means rules act together. In Fig.~\ref{fig:overview}(a,b), the KD example rotates commands by $90^\circ$ and rebounds on wall collision. Panel~(c) tests the four one-key KD combinations listed below.}
      \label{tab:training-test-settings}
\normalsize
\setlength{\tabcolsep}{5pt}
\renewcommand{\arraystretch}{1.12}
\renewcommand{\tabularxcolumn}[1]{m{#1}}
\setlength{\defaultaddspace}{6pt}
\begin{tabularx}{\linewidth}{@{}>{\raggedright\arraybackslash}m{.16\linewidth}>{\hsize=.92\hsize\linewidth=\hsize\raggedright\arraybackslash}X>{\hsize=1.08\hsize\linewidth=\hsize\raggedright\arraybackslash}X@{}}
\toprule
\textbf{Rule family} & \textbf{Train rules} & \textbf{Test rules} \\
\midrule
\multicolumn{3}{@{}l}{\textbf{MechanismGrid} \quad Atomic rules and unseen pairs (Fig.~\ref{fig:icl-prediction}a)} \\
\addlinespace[2pt]
Single rules (9)
& One-cell move, $R_{90}$, $R_{180}$, $R_{270}$, south wind, column field, wall bounce, horizontal reflection, even-row gate.
& \emph{Familiar-rule control.} The nine training rules on new tasks. \\
\addlinespace
Held-out pairs (10) & \textemdash
& $R_{90}$ + south wind, $R_{180}$ + south wind, \mbox{$R_{270}$ + south wind}\par
  $R_{90}$ + bounce, $R_{180}$ + bounce, $R_{270}$ + bounce\par
  column field + reflection, bounce + reflection\par
  south wind + row gate, reflection + row gate \\
\addlinespace
\midrule
\multicolumn{3}{@{}l}{\textbf{MechanismGrid} \quad Parameter holdout (Fig.~\ref{fig:icl-prediction}b)} \\
\multicolumn{3}{@{}p{\linewidth}@{}}{For each of the 13 targets, remove that rule from the training catalog below (42 source rules). For the column-field target, add all four split fields with two-cell wind (46 source rules).} \\
\addlinespace
Move length & 1, 2 or 3 cells & 1, 2 or 3 cells \\
\addlinespace[3pt]
Direction & 23 nonidentity permutations of the four directions & $R_{90}$, $R_{180}$, $R_{270}$, horizontal reflection \\
\addlinespace
Wind & 1 or 2 cells north, south, east or west & 1 cell north, south, east or west \\
\addlinespace[3pt]
Wind field & Row- and column-split fields, both orientations & Column split (south on left, north on right) \\
\addlinespace
Gate & Row and column gates, each on odd or even indices & Even-row gate (blocks horizontal moves) \\
\addlinespace
Collision & Wall bounce & \textemdash \\
\midrule
\multicolumn{3}{@{}l}{\textbf{KeyDoor} \quad One key, held-out motion combinations} \\
\addlinespace[2pt]
Single rules & One-cell move, $R_{90}$, $R_{180}$, $R_{270}$, horizontal reflection, wall bounce & \textemdash \\
\addlinespace
Pairs & $R_{90}$ + reflection, $R_{270}$ + reflection\par
        $R_{90}$ + bounce, $R_{270}$ + bounce
      & $R_{180}$ + reflection, $R_{180}$ + bounce\par
        reflection + bounce \\
\addlinespace
Triples & $R_{90}$ + reflection + bounce\par $R_{270}$ + reflection + bounce
        & $R_{180}$ + reflection + bounce \\
\addlinespace
\midrule
\multicolumn{3}{@{}l}{\textbf{Continuous2D} \quad Parameter holdout by family} \\
\addlinespace[2pt]
Horizontal scale & Identity and factors $0.60, 0.70, 0.80, 0.90$, $1.10, 1.15, 1.20, 1.25$ & $0.50, 1.40$ \\
\addlinespace
Drift & Identity, or $0.10$ or $0.15$ in each cardinal direction & North $0.25$, east $0.25$ \\
\addlinespace
Right-half rotation & Identity and $\pm10^\circ, \pm20^\circ, \pm30^\circ, \pm40^\circ$ & $-60^\circ, +60^\circ$ \\
\midrule
\textbf{PointMass} & Force gains $-3,-2,+2,+3$ & Force gains $-4,+4$ \\
\addlinespace[3pt]
\textbf{Reacher} & Actuator gains $-3,-2,+2,+3$ & Actuator gains $-4,+4$ \\
\bottomrule
\end{tabularx}

    \end{minipage}
  \end{lrbox}

\begin{AppendixProtocols}
  \section{Models, Objectives, and Training}
  \label{app:training}
  A world model (WM) predicts the next image embedding. A separate distance head uses the trained WM for navigation.

  \paragraph{World model.}
  Each ICL WM uses a trainable CNN with 128-dimensional image embeddings and one eight-head cross-attention layer. Each context token represents an action and its before/after images. The query attends to all $K$ tokens and predicts a change from the current embedding. Context tokens have no position index, and queries are processed independently. The query successor supplies the training target. Scaling varies network widths and keeps 128-dimensional embeddings and one attention layer.

  \paragraph{Objectives.}
  MG and C2D minimize the mean squared error (MSE) between predicted and encoded successor images, plus SIGReg with weight 0.09 (Eq.~\ref{eq:wm-loss}). SIGReg regularizes the observed image embeddings. PointMass and Reacher average errors over two steps, feeding the first prediction into the second step.

  \paragraph{KD auxiliary objective.}
  KD adds $0.1(L_{\rm pos}+0.4L_{\rm inv}+0.5\sum_j L_{{\rm key},j})$ to MSE plus SIGReg. Linear classifiers read the predicted successor embedding and predict the agent cell, the set of held keys, and each key cell or held status. Labels come from the true successor pixels. These cross-entropies train both the classifiers and the WM. KD scaling selects models by the full loss. KD ICL evaluation loss includes MSE and SIGReg. State accuracy and navigation use the WM without the classifiers.

  \paragraph{Optimization and freezing.}
  WM training uses AdamW at $3\times10^{-4}$ with 64 queries per update. MG ICL samples integer $K$ from 0 to 256, and MG scaling samples from $\{0,64,256\}$. Other studies sample from $\{0,1,2,4,8,16,32,64,128,256\}$. ICL uses val-selected checkpoints or fixed training endpoints. Scaling uses fixed compute endpoints. We train the distance head with the WM fixed. Both sets of weights stay fixed at test time.
\end{AppendixProtocols}
\FloatBarrier
\newsavebox{\AppendixPageBox}
\newlength{\AppendixColumnWidth}
\setlength{\AppendixColumnWidth}{\columnwidth}
\newcommand{\AppendixPage}[2]{%
  \sbox{\AppendixPageBox}{\begin{minipage}{\textwidth}\raggedbottom #2\end{minipage}}%
  \typeout{APPENDIX-PAGE #1 HEIGHT=\the\dimexpr\ht\AppendixPageBox+\dp\AppendixPageBox\relax LIMIT=\the\textheight}%
  \ifdim\dimexpr\ht\AppendixPageBox+\dp\AppendixPageBox\relax>\dimexpr\textheight-5pt\relax
    \PackageWarning{AppendixLayout}{Page group #1 exceeds the available height}%
  \fi
  \clearpage
  \begingroup
  \ifdefined\nolinenumbers\nolinenumbers\fi
  \noindent\usebox{\AppendixPageBox}\par
  \endgroup
}
\newcommand{\AppendixColumns}[2]{%
  \par\noindent\raisebox{-\height}{\begin{minipage}[t]{\AppendixColumnWidth}#1\end{minipage}}\hfill
  \raisebox{-\height}{\begin{minipage}[t]{\AppendixColumnWidth}#2\end{minipage}}\par
}
\newcommand{\AppendixText}[1]{%
  \begingroup
  \setlength{\multicolsep}{4pt plus 1pt minus 1pt}%
  \begin{multicols}{2}\raggedcolumns #1\end{multicols}%
  \endgroup
}
\newcommand{\AppendixFigure}[4][1]{%
  \par\begingroup\centering\includegraphics[width=#1\linewidth]{figures/#2}\par
  \captionsetup{hypcap=false}\captionof{figure}{#3}\label{#4}\def\AppActualLabel{#4}\def\AppMatrixAlias{fig:mg-strict-prediction}\ifx\AppActualLabel\AppMatrixAlias\label{fig:validation-loss-heatmap}\fi\endgroup\par%
}
\newcommand{\AppendixTable}[3]{%
  \par\begingroup\centering\footnotesize\setlength{\tabcolsep}{3pt}
  \captionsetup{hypcap=false}\captionof{table}{#2}\label{#3}
  \input{tables/#1}\par\endgroup\medskip%
}
\raggedbottom

\onecolumn
\begingroup
\renewcommand{\thefigure}{\getrefnumber{app:environments}.\arabic{figure}}
\AppendixPage{A-dynamics-mg}{
  \AppendixFigure{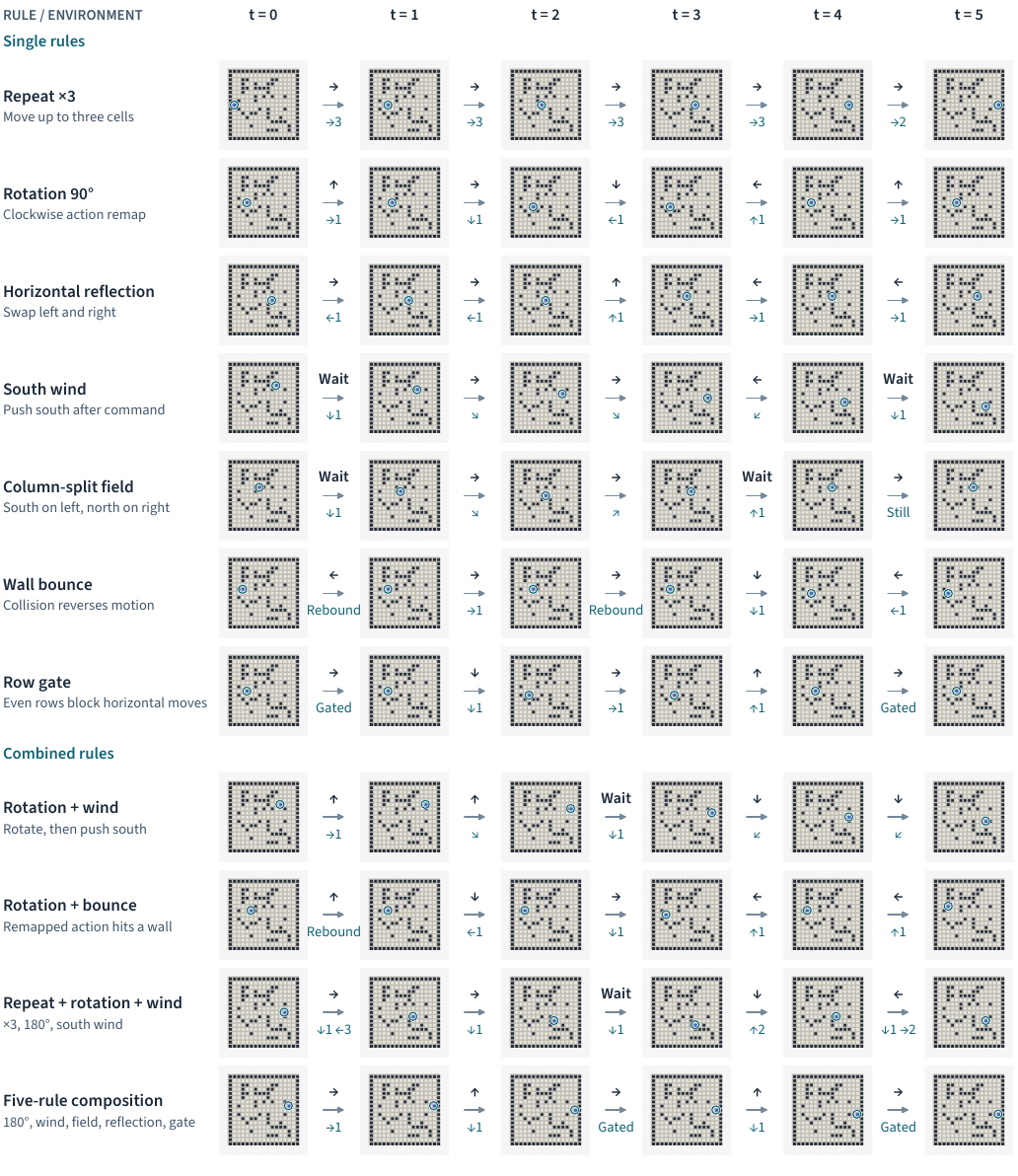}{%
    \textbf{How rules change movement in MechanismGrid.} Each row shows the initial RGB image and
    the result of five actions on one $17\times17$ map. Above each arrow is the commanded direction
    or wait. Below it is the observed movement or event. A diagonal arrow indicates one cell along
    each axis. The circle marks the blue agent.
  }{fig:mg-rgb-transitions}
}

\AppendixPage{A-dynamics-kd-continuous}{
  \AppendixFigure[.96]{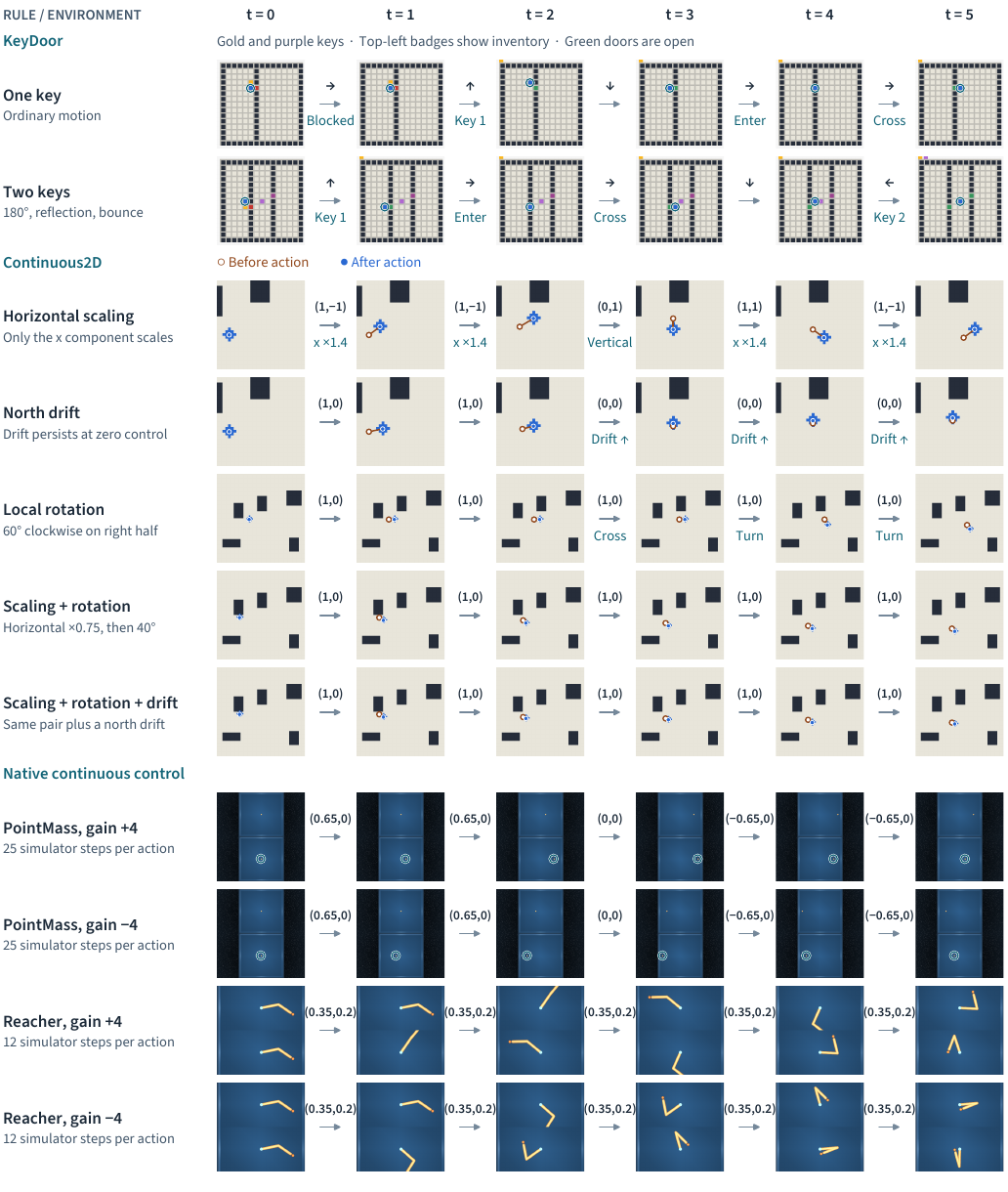}{%
    \textbf{Movement in KeyDoor, Continuous2D, PointMass, and Reacher.} Rows show initial RGB images
    and five actions. In KeyDoor, both keys start on the map in the two-key row. Collected keys
    appear as top-left badges, and matching doors turn green. In Continuous2D, orange rings mark
    starts and blue dots mark ends. The positive axes point right and down. Local rotation applies
    at starting $x\geq0.5$. The two C2D combinations share commands and starts. Within each
    simulator the two gains also share commands and starts. The upper frame shows the previous
    image.
  }{fig:kd-continuous-rgb-transitions}
}

\AppendixPage{A-dynamics-c2d-remaining}{
  \AppendixFigure{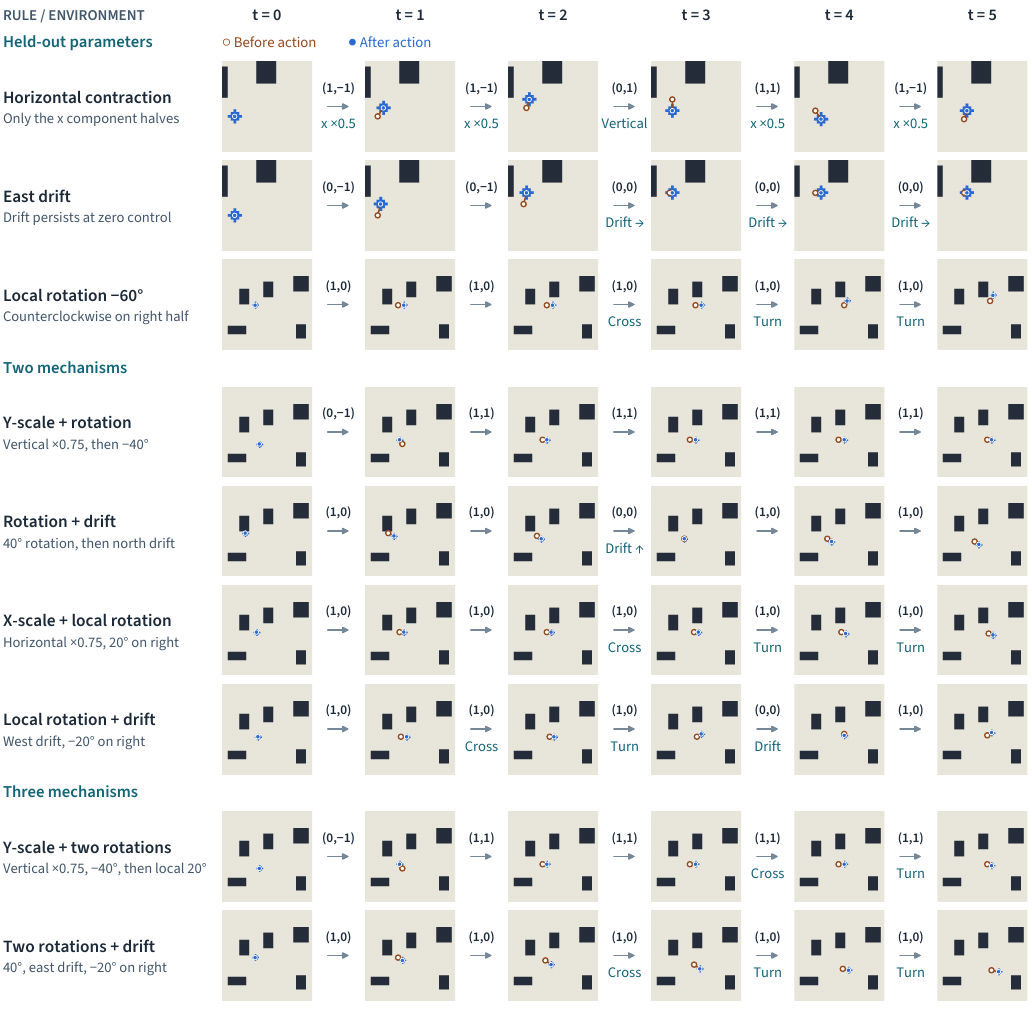}{%
    \textbf{More movement rules in Continuous2D.} Rows show three additional held-out parameter
    values, four pairs, and two triples. Together with Fig.~\ref{fig:kd-continuous-rgb-transitions},
    they illustrate all C2D test rules. Each arrow gives the action. The orange ring and blue dot
    mark the positions before and after it. Contraction and east drift use a fixed crop throughout
    the row. Local rotation applies at starting $x\geq0.5$.
  }{fig:c2d-remaining-rgb-transitions}
}
\endgroup

\AppendixPage{A-training-test-settings}{\usebox{\TrainingTestSettings}}

\AppendixPage{C1-scans}{
  \section{Scaling Evidence}\label{app:compute}
  \subsection{Trade-offs at Fixed Compute and Scaling Fits}\label{app:allocation-fits}
  \AppendixText{
    \paragraph{Parameters, data, and compute.}
    At fixed compute, larger WMs process fewer queries. $N$ counts WM parameters, excluding KD auxiliary classifiers. $D$ counts training queries, each paired with one context set. Different queries can describe the same physical transition. $C$ counts profiled forward and backward FLOPs, including the use of context and KD auxiliary prediction. It excludes optimizer updates, operators outside the profiler, distance-head training, and navigation.
  }
  \AppendixFigure{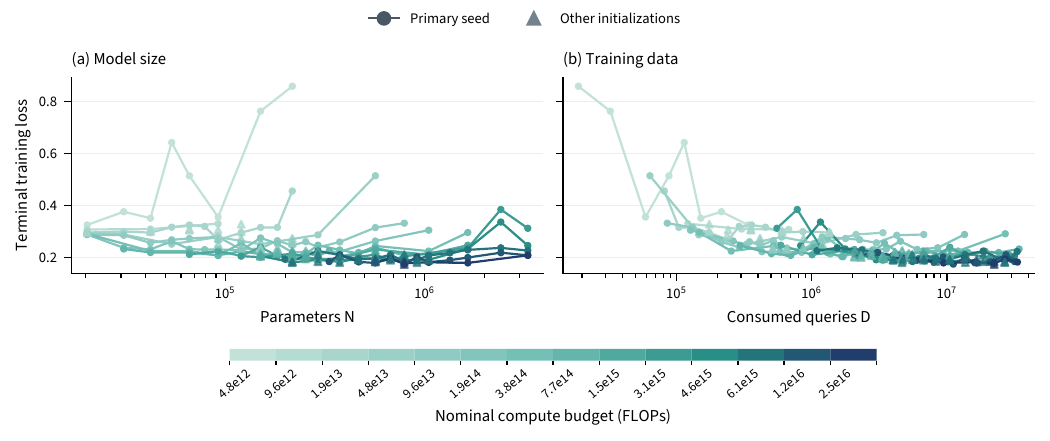}{%
    \textbf{MG training loss at fixed compute.} The left panel varies model size $N$, and the right
    varies training queries $D$. Each color shows one of the 14 nominal compute budgets used in the
    paper. Loss is averaged over the final 100 updates, or all updates for shorter runs. Circles
    show the main seed and triangles show other seeds. We show 223 runs and omit 19 with loss above
    twice the median at their budget.
  }{fig:mg-isoflop-all}
  \AppendixFigure[.985]{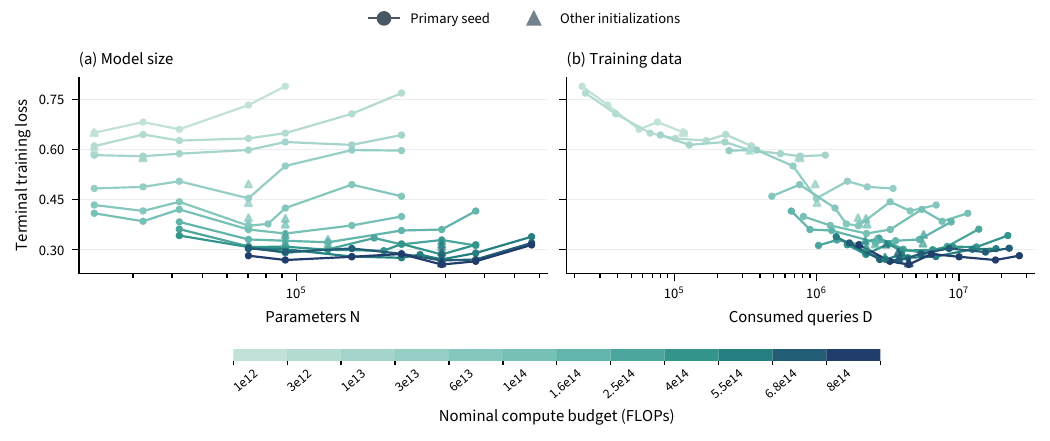}{%
    \textbf{KD training loss at fixed compute.} The same 110 runs appear against model size $N$
    (left) and training query count $D$ (right). Color shows one of 12 nominal compute budgets.
    Circles show the main seed and triangles show other seeds. Final training loss includes the
    RGB-supervised key and position terms. Two runs with loss above twice the median at their budget
    are omitted.
  }{fig:kd-isoflop-all}
}

\AppendixPage{C1-selection}{
\subsubsection{Selected checkpoints and descriptive fits}
  \par\begingroup\centering\footnotesize\setlength{\tabcolsep}{3pt}
  \captionsetup{hypcap=false}\captionof{table}{%
  \textbf{Lowest-training-loss model at each compute budget.} We select one of the tested $(N,D)$
  allocations from seed 2026092880 at each of 14 MG and 12 KD budgets. $N$ is model size, $D$ is the
  query count, and $L^*$ is mean loss over the final 100 updates (or all updates for shorter runs).
  $C$ is the actual FLOP count, rounded in the table.
}\label{tab:strict-models}
  \centering
\begin{tabular}{@{}rrrrr@{\hspace{1.7em}}rrrrr@{}}
\toprule
\multicolumn{5}{c}{MG} & \multicolumn{5}{c}{KD} \\
\cmidrule(r){1-5}\cmidrule(l){6-10}
$C$ (FLOPs) & $N$ & $D$ & Updates & $L^*$ & $C$ (FLOPs) & $N$ & $D$ & Updates & $L^*$ \\
\midrule
4.80e12 & 20,320 & 338,944 & 5,296 & 0.3253 & 9.99e11 & 22,548 & 117,632 & 1,838 & 0.6504 \\
9.60e12 & 20,320 & 674,560 & 10,540 & 0.3079 & 3.00e12 & 22,548 & 350,016 & 5,469 & 0.6102 \\
1.92e13 & 92,176 & 242,176 & 3,784 & 0.2677 & 1.00e13 & 32,311 & 769,856 & 12,029 & 0.5801 \\
4.80e13 & 150,440 & 329,792 & 5,153 & 0.2512 & 3.00e13 & 70,236 & 996,160 & 15,565 & 0.4540 \\
9.60e13 & 120,272 & 869,760 & 13,590 & 0.2310 & 6.00e13 & 70,236 & 1,971,072 & 30,798 & 0.3722 \\
1.92e14 & 216,992 & 841,600 & 13,150 & 0.2199 & 1.00e14 & 92,176 & 2,241,856 & 35,029 & 0.3487 \\
3.84e14 & 216,992 & 1,664,448 & 26,007 & 0.2039 & 1.60e14 & 126,212 & 2,597,248 & 40,582 & 0.3225 \\
7.68e14 & 216,992 & 3,326,784 & 51,981 & 0.1996 & 2.50e14 & 92,176 & 5,619,776 & 87,809 & 0.2906 \\
1.54e15 & 291,832 & 4,616,768 & 72,137 & 0.1919 & 4.00e14 & 291,832 & 2,218,496 & 34,664 & 0.2861 \\
3.07e15 & 374,960 & 6,813,120 & 106,455 & 0.1868 & 5.50e14 & 291,832 & 3,050,368 & 47,662 & 0.2712 \\
4.61e15 & 216,992 & 20,006,144 & 312,596 & 0.1813 & 6.80e14 & 291,832 & 3,769,408 & 58,897 & 0.2681 \\
6.14e15 & 790,352 & 5,556,928 & 86,827 & 0.1846 & 8.00e14 & 291,832 & 4,436,288 & 69,317 & 0.2559 \\
1.23e16 & 790,352 & 11,131,648 & 173,932 & 0.1745 &  &  &  &  &  \\
2.46e16 & 790,352 & 22,280,768 & 348,137 & 0.1780 &  &  &  &  &  \\
\bottomrule
\end{tabular}
\par\endgroup\medskip%

\AppendixColumns{
\paragraph{Selecting models.}
At each compute budget, we compare the tested $(N,D)$ configurations trained with seed 2026092880. We select the one with the lowest mean training loss over its final 100 updates (or all updates if fewer), breaking ties by lower $N$, then lower $D$. Only the selected WMs receive the full split/context evaluation.

\paragraph{Fitting the selected points.}
Using the unrounded actual FLOPs in Table~\ref{tab:strict-models}, we separately fit each $Y\in\{N^*,D^*,L^*\}$ by equally weighted least squares in log space, with $Y=Ax^b$ and $x=C/10^{14}$. Nearby allocations can have almost the same loss. A small loss difference can then change $N^*$ and $D^*$ substantially while barely changing $L^*$. Table~\ref{tab:scaling-fit-errors} shows larger percentage errors for $N^*$ and $D^*$ in both environments.
\begin{align*}
N^*_{\mathrm{MG}}(C)&\approx 1.0681\times10^{5}x^{0.39199},\\
D^*_{\mathrm{MG}}(C)&\approx 1.0090\times10^{6}x^{0.52996},\\
L^*_{\mathrm{MG}}(C)&\approx 0.24164x^{-0.07242}.\\[2pt]
N^*_{\mathrm{KD}}(C)&\approx 1.0961\times10^{5}x^{0.42734},\\
D^*_{\mathrm{KD}}(C)&\approx 1.8272\times10^{6}x^{0.49036},\\
L^*_{\mathrm{KD}}(C)&\approx 0.35391x^{-0.15356}.
\end{align*}

}{
  \par\begingroup\centering\footnotesize\setlength{\tabcolsep}{3pt}
  \captionsetup{hypcap=false}\captionof{table}{%
  \textbf{Fit error across compute budgets.} The first column fits and scores on all budgets. The
  second refits without each budget and scores that budget. Both columns average the absolute
  percentage error for selected model size $N^*$, query count $D^*$, or training loss $L^*$.
}\label{tab:scaling-fit-errors}
  \begin{tabular}{@{}llrr@{}}
\toprule
\multirow{2}{*}{Environment} & \multirow{2}{*}{Quantity} & \multicolumn{2}{c}{MAPE (\%, lower is better)} \\
 & & All budgets & Leave one out \\
\midrule
\multirow{3}{*}{\MGabbr} & $N^*$ & 37.68 & 47.53 \\
 & $D^*$ & 40.47 & 47.86 \\
 & $L^*$ & 4.19 & 4.99 \\
\midrule
\multirow{3}{*}{\KDabbr} & $N^*$ & 23.20 & 28.01 \\
 & $D^*$ & 28.29 & 38.13 \\
 & $L^*$ & 3.76 & 4.90 \\
\bottomrule
\end{tabular}
\par\endgroup\medskip%

\subsubsection{Compute per example}\label{app:compute-cost}
With batch size $B$ and $S$ updates, $D=BS$ and $C\approx D f(N)$, where $f(N)$ is the profiled forward and backward FLOPs per query. Averaging by model size gives $f(N)\approx q_0(N/10^5)^{1.2}$, with $q_0\approx9\times10^7$ (MG) and $5\times10^7$ (KD). The separate exponents are 1.204 and 1.171 over 20 and 13 sizes, respectively. On the selected allocations, the rounded formula has mean absolute relative errors of 2.4\% in MG and 2.7\% in KD.

}
\AppendixColumns{
\subsubsection{Checking loss forecasts}
Table~\ref{tab:retrospective-loss-checks} checks loss forecasts in a separate set of already trained WMs. MG uses 28 models and nine test context sizes (all except 128). KD uses eight models, one train loss averaged over mixed $K$ per model, and test losses at $K=0,64,256$.
}{
  \par\begingroup\centering\footnotesize\setlength{\tabcolsep}{3pt}
  \captionsetup{hypcap=false}\captionof{table}{%
  \textbf{Error in forecasts of total WM loss.} Each of the $n$ pairs compares predicted and
  measured loss for the same model and context setting. APE is absolute percentage error,
  $100|L_{\rm pred}-L_{\rm obs}|/L_{\rm obs}$.
}\label{tab:retrospective-loss-checks}
  \begin{tabular}{@{}lrrr@{}}
\toprule
Loss readout & Loss pairs $n$ & Mean APE & Median APE \\
\midrule
MG test (nine $K$) & 252 & 15.3\% & 14.6\% \\
KD train (mixed $K$) & 8 & 7.6\% & 6.3\% \\
KD test (three $K$) & 24 & 7.3\% & 4.6\% \\
\bottomrule
\end{tabular}
\par\endgroup\medskip%

}
}

\AppendixPage{C2-MG-prediction}{
  \subsection{Prediction by Split and Compute}\label{app:strict-prediction}\label{app:validation-grid}
  \AppendixText{
    Figures~\ref{fig:mg-strict-prediction} and~\ref{fig:kd-strict-prediction} evaluate the models in Table~\ref{tab:strict-models}. For train, each WM is scored on 128 batches replayed from its training run. The val queries and the test queries are each shared across budgets within an environment. Every split uses the same 8,192 queries at every $K$. Accuracy uses the decoded agent position in MG and both position and held keys in KD (Appendix~\ref{app:metrics}). Figure~\ref{fig:mg-prediction-accuracy} shows the same MG test accuracy.
  }
  \AppendixFigure{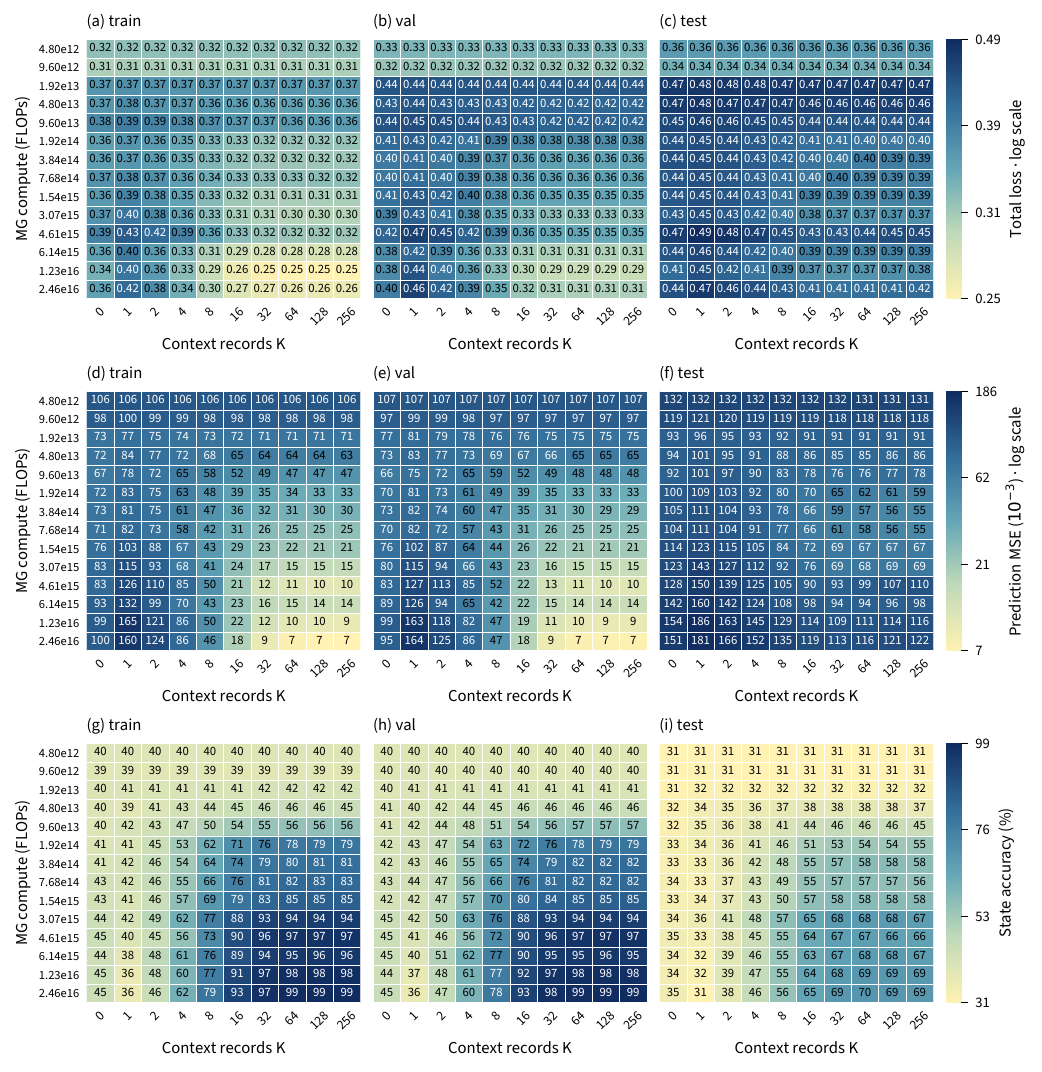}{%
    \textbf{MG prediction on train, val, and test.} Columns show the three splits. Panel rows give
    MSE plus SIGReg, next-image embedding MSE (in $10^{-3}$ units), and agent-cell accuracy (\%).
    Each cell averages 8,192 queries at one compute budget and context size $K$ (transition
    records). We use the lowest-training-loss model at each of 14 budgets. Scales are shared across
    splits (log for losses and linear for accuracy).
  }{fig:mg-strict-prediction}
}

\AppendixPage{C2-KD-prediction}{
  \AppendixFigure{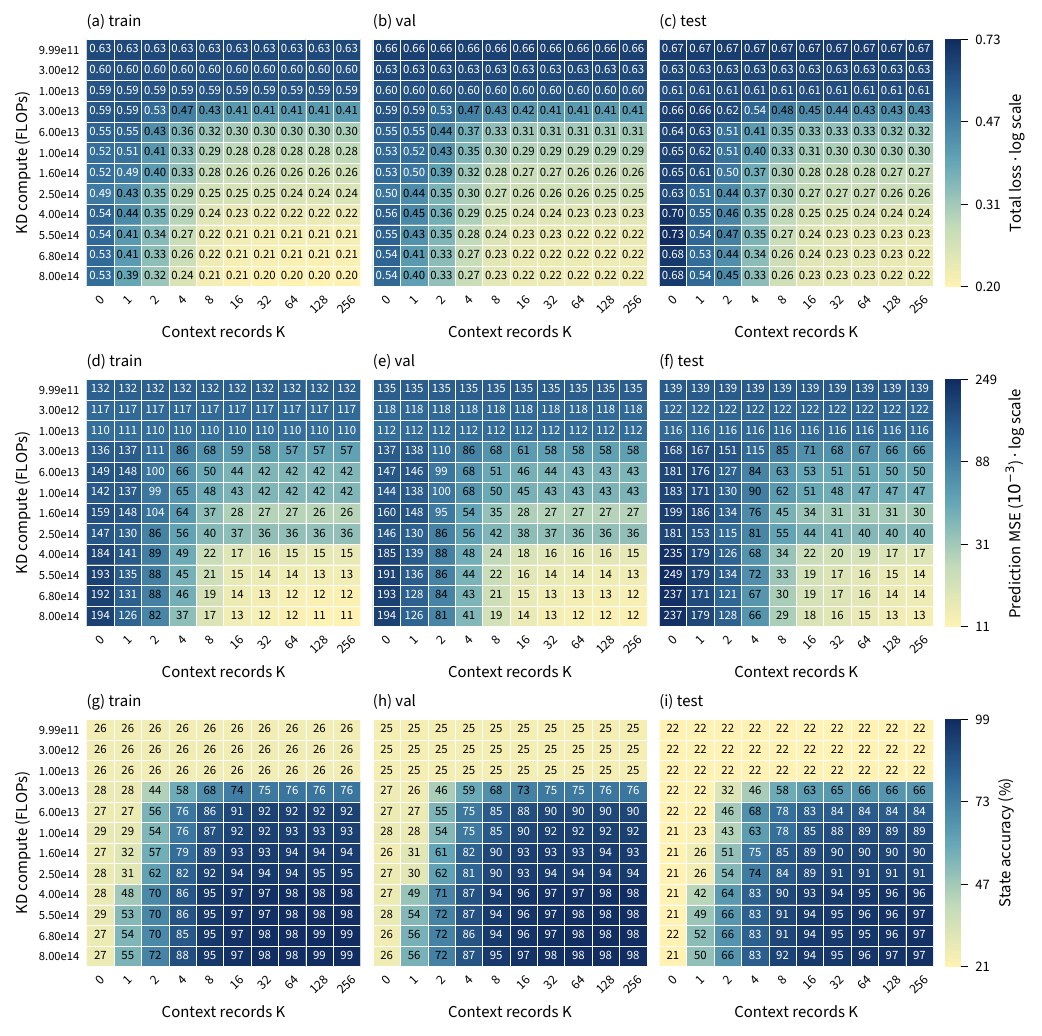}{%
    \textbf{KD prediction on train, val, and test.} Columns show the three splits. Panel rows give
    MSE plus SIGReg and the KD auxiliary loss, next-image embedding MSE (in $10^{-3}$ units), and
    joint position-and-inventory accuracy (\%). Each cell averages 8,192 queries at one compute
    budget and context size $K$ (transition records). We use the lowest-training-loss model at each
    of 12 budgets. Scales are shared across splits (log for losses and linear for accuracy).
  }{fig:kd-strict-prediction}
  \AppendixColumns{
    \paragraph{Reading the metrics together.}
    Total loss combines MSE and SIGReg, plus KD's auxiliary. Accuracy measures the predicted complete state. Changing encoders can change latent scale, so lower MSE across WMs is not by itself evidence of more accurate state prediction.
  }{
    \paragraph{MG error at larger budgets.}
    At large $K$, MG train MSE falls overall as compute rises. At high compute, test MSE rises even as accuracy stays near its peak. Figure~\ref{fig:mg-mse-training} tracks MSE during training at fixed model size.
  }
}

\AppendixPage{C3-navigation}{
  \subsection{Navigation by Compute and Context}\label{app:primary-compute-navigation}
  \AppendixText{
    Figures~\ref{fig:mg-primary-navigation} and~\ref{fig:kd-primary-navigation} extend Fig.~\ref{fig:navigation-compute-control} to ten context sizes. At each budget, the planners share the same WM, test tasks, and context records. Appendix~\ref{app:control} explains how each planner scores actions.
  }
  \AppendixFigure{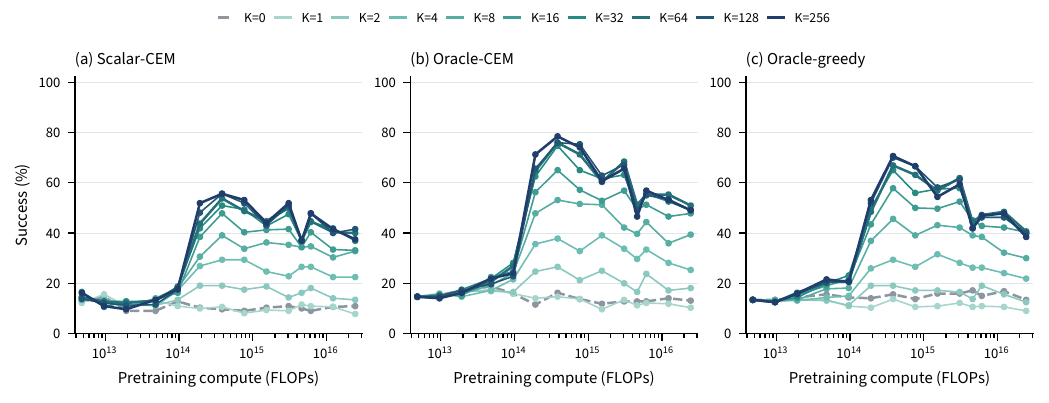}{%
    \textbf{MG navigation on ten rule pairs withheld from training.} Panels show learned-distance
    Scalar-CEM, Oracle-CEM with decoded-state distances, and one-step Oracle-greedy. Each point
    shows the percentage of the same 320 test tasks (32 per pair) solved within 60 actions. The
    horizontal axis gives actual pretraining FLOPs for the 14 models used in
    Fig.~\ref{fig:navigation-compute-control}(a). Colors mark the number $K$ of fixed context
    records. The dashed line shows $K=0$ (no context).
  }{fig:mg-primary-navigation}
  \AppendixFigure{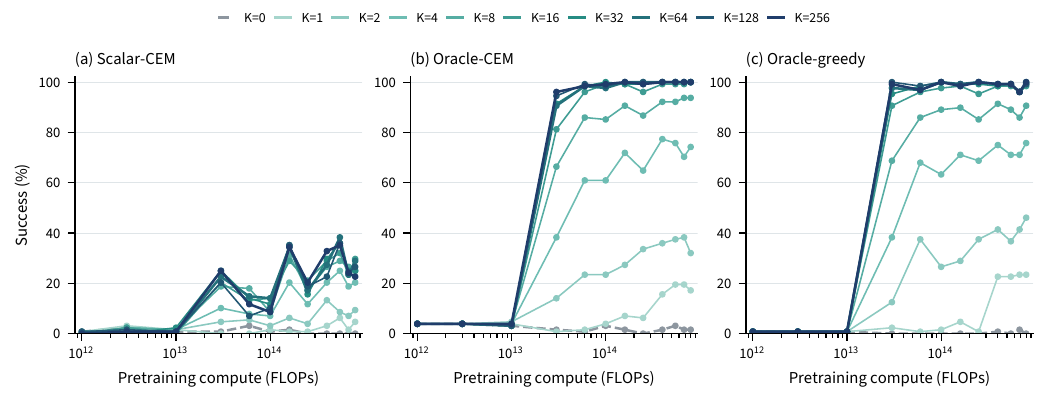}{%
    \textbf{KD goal-reaching success on unseen rule combinations with one key.} Panels show
    learned-distance Scalar-CEM, Oracle-CEM with decoded-state distances, and one-step Oracle-greedy
    at 12 pretraining compute budgets. Each point is the percentage of 128 test tasks solved within
    256 actions. Colors show the ten fixed context sizes $K$. The dashed line uses no context. All
    planners at a budget use the same WM as Fig.~\ref{fig:kd-strict-prediction}.
  }{fig:kd-primary-navigation}
}

\AppendixPage{C4-primary-context-benefit}{
  \subsection{Accuracy and MSE Changes from No Context}\label{app:prediction-confirmation}
  \AppendixText{
    At each budget, we use one trained WM and the same 8,192 test queries at each $K$. Every curve in Figs.~\ref{fig:mg-compute-prediction} and~\ref{fig:kd-compute-prediction} compares a context size with the same model at $K=0$ (Eq.~\ref{eq:gain}).
  }
  \AppendixFigure{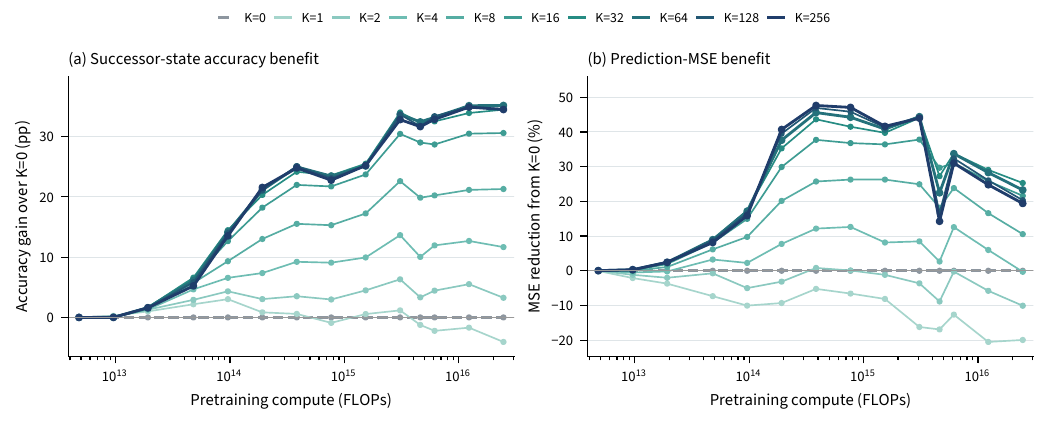}{%
    \textbf{MG prediction gains from context on unseen test pairs.} Across 14 pretraining compute
    budgets, (a) shows the change in agent-cell accuracy in percentage points relative to $K=0$. In
    (b), $100(1-\mathrm{MSE}_K/\mathrm{MSE}_0)$ gives the percentage reduction in next-image
    embedding MSE. Positive values mean higher accuracy or lower MSE. Each color is a context size
    $K$, using the same 8,192 test queries at each size. At the largest budget, $K=256$ raises
    accuracy by 34.5 percentage points.
  }{fig:mg-compute-prediction}
  \AppendixFigure{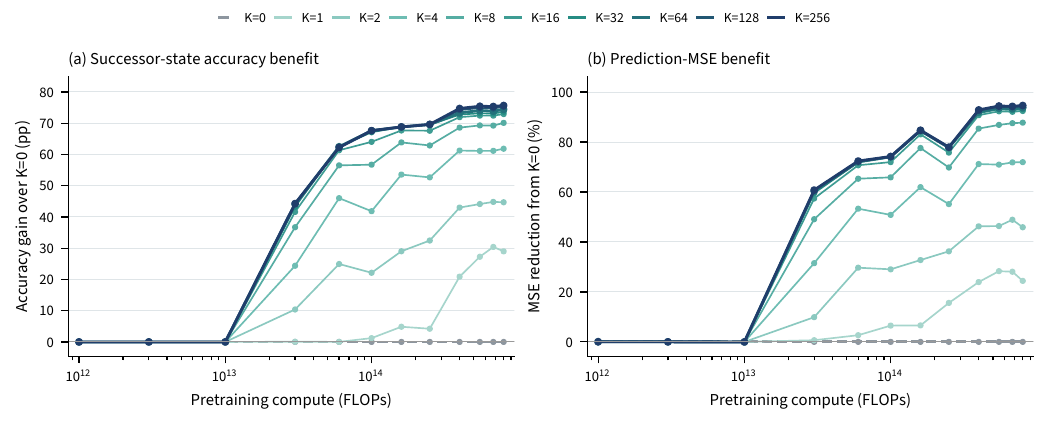}{%
    \textbf{KD prediction gains from context on unseen one-key test combinations.} Across 12
    pretraining compute budgets, (a) shows the change in joint position-and-inventory accuracy in
    percentage points relative to $K=0$. In (b), $100(1-\mathrm{MSE}_K/\mathrm{MSE}_0)$ gives the
    percentage reduction in next-image embedding MSE. Positive values mean higher accuracy or lower
    MSE. Each color is a context size $K$, using the same 8,192 test queries at each size. At the
    largest budget, $K=256$ raises accuracy by 75.7 percentage points.
  }{fig:kd-compute-prediction}
  \AppendixText{
    In separate three-seed checks, the $K=256$ accuracy gain rises at each of four MG budgets on new maps and at each of three KD budgets on new episodes from previously used layouts. Two further MG budgets reuse the confirmation maps. On those maps the gain falls at the last budget for two of three seeds.
  }
}

\AppendixPage{C5-MSE-trajectories}{
  \subsection{Prediction Error during Training}\label{app:development-dynamics}
  \AppendixColumns{
    We train two WMs for each of six configurations separate from Table~\ref{tab:strict-models}. Each pair shares the model size, training queries, stopping step, seed, and sampling order, with SIGReg weight set to 0.09 or 0.01. Each measurement averages 2,048 fixed queries (32 train layouts and 77 layouts each for val and test). At $K=256$ and weight 0.09, late MSE falls on train and val in all six runs and rises on test in five (Table~\ref{tab:late-mse}).
  }{
    \renewcommand{\arraystretch}{.95}%
  \par\begingroup\centering\footnotesize\setlength{\tabcolsep}{3pt}
  \captionsetup{hypcap=false}\captionof{table}{\textbf{Change in MG prediction MSE late in training.} Percent change from mean MSE at 50--75\% of updates to 75--100\% of updates. Results use fixed queries, $K=256$, SIGReg weight 0.09, and one seed.}\label{tab:late-mse}
  \begin{tabular}{@{}rrrrr@{}}
\toprule
$C$ (FLOPs) & $N$ & train & val & test \\
\midrule
7.68e14 & 150,440 & -7.5\% & -7.2\% & -2.3\% \\
1.54e15 & 216,992 & -11.2\% & -10.2\% & +6.8\% \\
1.54e15 & 253,376 & -10.1\% & -13.2\% & +3.1\% \\
3.07e15 & 291,832 & -16.2\% & -19.7\% & +7.5\% \\
4.61e15 & 374,960 & -27.4\% & -27.1\% & +22.0\% \\
6.14e15 & 466,376 & -29.4\% & -30.7\% & +15.0\% \\
\bottomrule
\end{tabular}
\par\endgroup\medskip%

  }
  \AppendixFigure[.925]{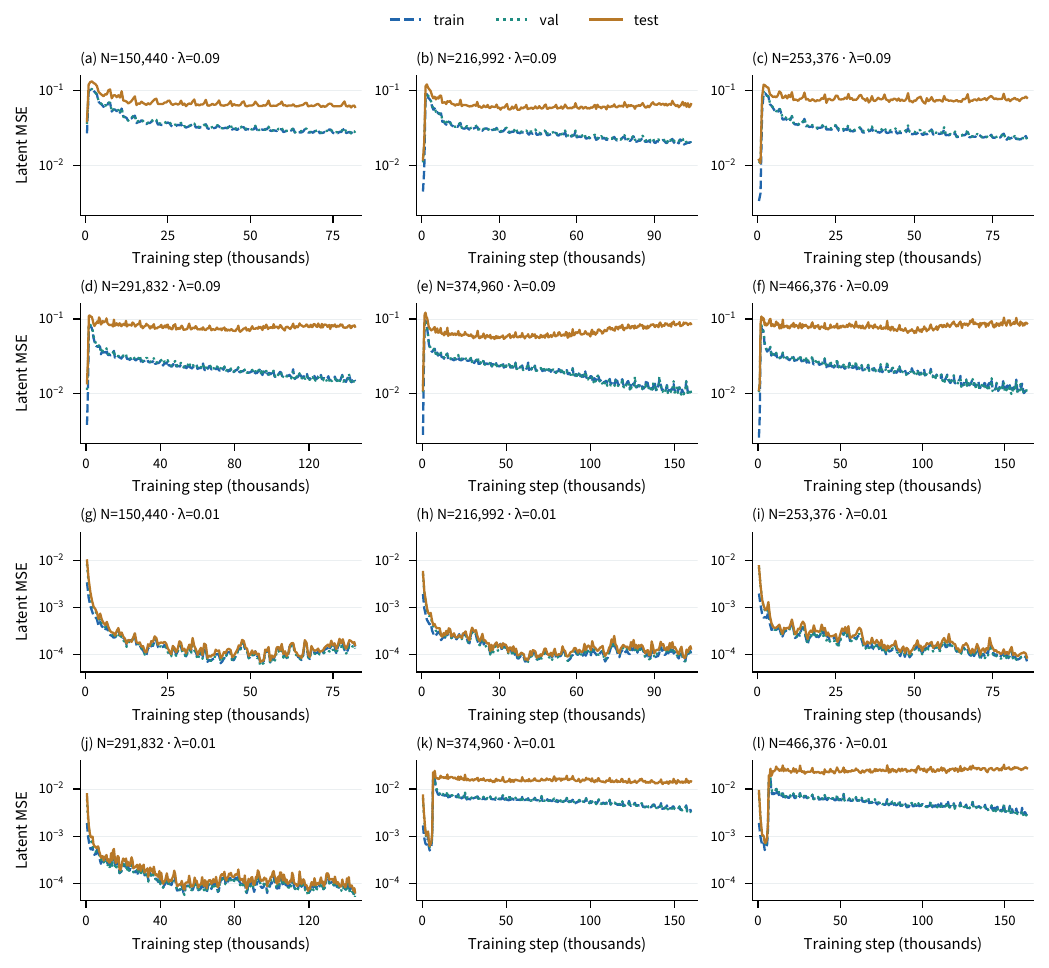}{%
    \textbf{MG next-image embedding MSE during training.} The upper six panels use SIGReg weight
    0.09 and the lower six use 0.01, with the same six model sizes. Lines use fixed queries per
    split at $K=256$. Blue dashes show train, teal dots show val, and solid ochre shows test. Panels
    at the same SIGReg weight share a log MSE scale. The 12 runs contribute 2,964 checkpoints.
  }{fig:mg-mse-training}
}

\AppendixPage{C5-objective-trajectories}{
  \subsubsection*{Total loss during training}
  \AppendixText{
    Total loss can fall as MSE rises if SIGReg falls by more. Figures~\ref{fig:mg-online-development-progress} and~\ref{fig:mg-low-sigreg-development-progress} show the total loss for the corresponding runs from Fig.~\ref{fig:mg-mse-training}.
  }
  \AppendixFigure[.985]{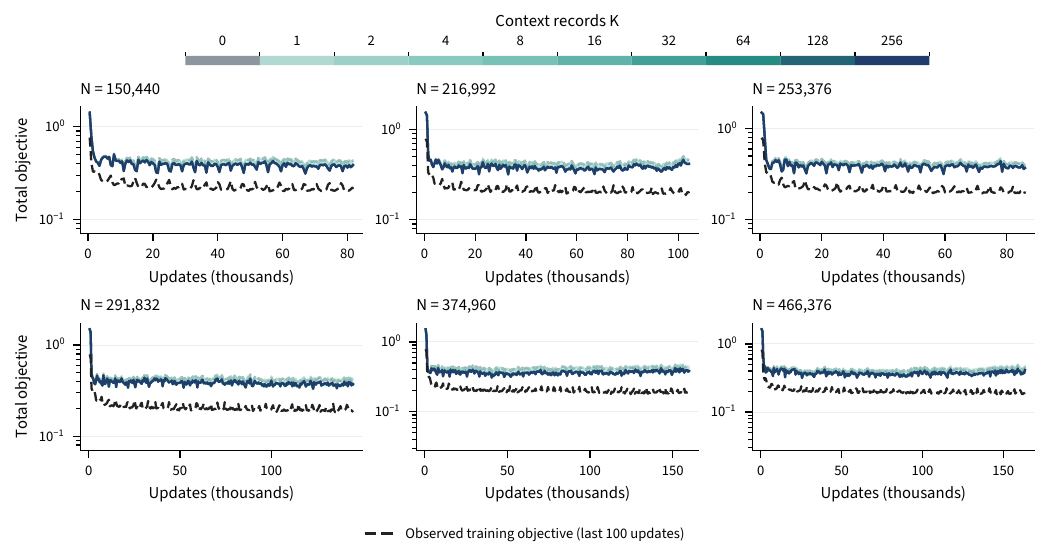}{%
    \textbf{MG total loss during training at SIGReg weight 0.09.} Each panel fixes one model size.
    Colored curves show MSE plus weighted SIGReg at ten fixed context sizes on the same test
    queries. The dashed black curve averages the last 100 training updates, which use a mix of
    $K=0,64,256$. The horizontal axis counts updates in thousands.
  }{fig:mg-online-development-progress}
  \AppendixFigure[.985]{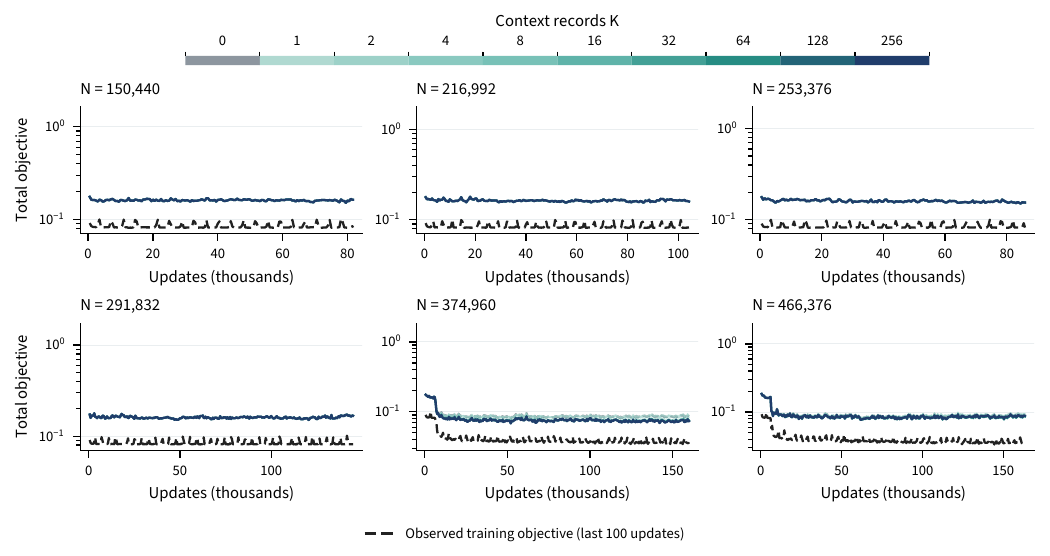}{%
    \textbf{MG total loss during training at SIGReg weight 0.01.} Model sizes, training steps, and
    test queries match Fig.~\ref{fig:mg-online-development-progress}. Colors show test loss for ten
    fixed context sizes. The dashed black curve averages the last 100 training updates at mixed
    $K=0,64,256$.
  }{fig:mg-low-sigreg-development-progress}
}

\AppendixPage{C5-representations-and-D-context}{
  \AppendixColumns{
    \subsubsection*{When small errors are misleading}
    With SIGReg weight 0.01, the four smaller WMs produce very similar embeddings for different states on the same map. Their prediction error is small, and copying the current embedding does at least as well. Adding context changes little.

    \AppendixFigure{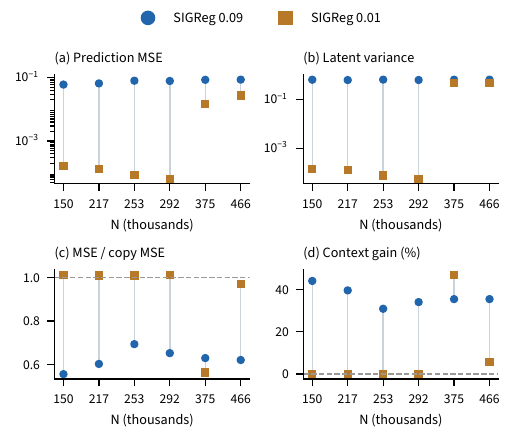}{%
      \textbf{MG embedding checks at the end of training.} Each point uses 2,048 test queries. At
      $K=256$, (a) gives prediction MSE, (b) shows target embedding variance within a map, and (c)
      divides prediction MSE by the error from copying the current embedding (below 1 favors the
      WM). In (d), $100(1-\mathrm{MSE}_{256}/\mathrm{MSE}_0)$ compares each WM with and without
      context.
    }{fig:mg-sigreg-representation}
  }{
    \section{Context Collection and Interventions}\label{app:context}
    \paragraph{How records are collected.}
    Each record contains an action and its before/after RGB images. MG, C2D, PointMass, and Reacher interleave sixteen 16-step random walks, taking one step per walk in each round. KD uses one 256-step walk from the initial position with no keys. Actions are uniform, including each continuous component in $[-1,1]$. We keep blocked moves and use the first $K$ records. Static context stays fixed. Online navigation adds each new transition.

    \paragraph{What the interventions test.}
    Table~\ref{tab:interventions} uses three WMs and ten unseen MG pairs (32 test maps per pair). Models, queries, tasks, and planning seeds stay fixed across conditions. Nonempty contexts have 256 records. Tasks allow 28 actions, compared with 45 in Fig.~\ref{fig:mg-pair-details}. Figure~\ref{fig:context-relations-current} tests prediction on the separate MG parameter-holdout WMs.
  }
  \AppendixFigure{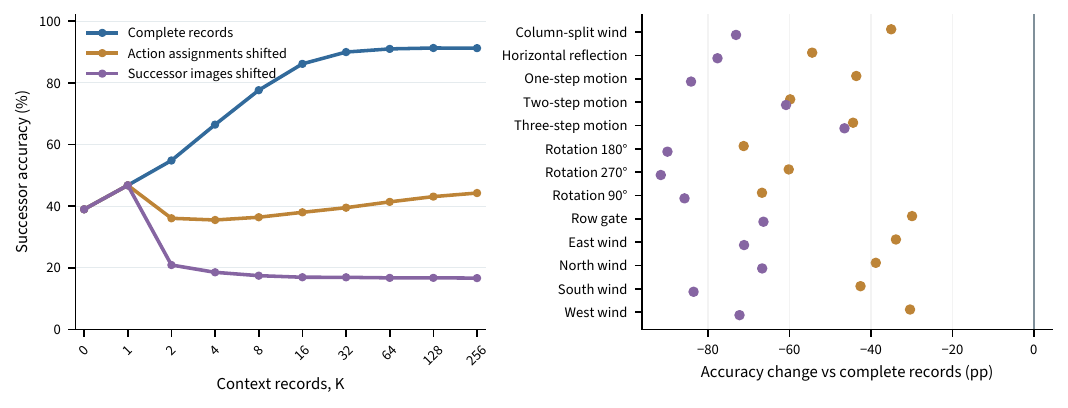}{%
    \textbf{MG prediction after reassigning actions or outcomes across context records.} Each
    control moves one field to the next record, wrapping the last to the first. Contexts of size
    zero or one are unchanged. The left panel averages test agent-cell accuracy over 13 held-out
    parameters, with three WMs per target. The right shows the difference from intact records for
    each target at $K=256$, in percentage points. Negative values mean worse accuracy.
  }{fig:context-relations-current}
}

\twocolumn
\begin{AppendixProtocols}
  \section{Metrics and Uncertainty}\label{app:metrics}
  Prediction is measured per query and goal reaching per episode. Variation is measured across independently trained models.

  \paragraph{Successor accuracy.}
  The evaluator embeds candidate successor images and retrieves the one closest to the WM prediction in Euclidean distance. MG candidates cover the agent positions in free cells. KD candidates cover reachable position and inventory states. A prediction counts as correct when the decoded cell and, in KD, all held-key bits match the true successor. KD auxiliary classifiers are used only in training.

  For C2D we retrieve from 64 candidate successor images. A match requires the position to agree after rounding to the rendering grid. For PointMass and Reacher, candidates share the current frame and differ in the future frame. Matching the correct candidate therefore depends on the future frame. Only the evaluator uses candidate labels.

  \paragraph{Prediction loss.}
  Prediction MSE is the mean squared difference between the predicted embedding and the encoding of the true next image. Different WMs can use different embedding scales, so we also report accuracy. Scaling evaluations reuse the same queries and random SIGReg projections across $K$. Some older ICL loss measurements do not use this pairing.

  \paragraph{Goal reaching and path efficiency.}
  Success rate (SR) is the percentage of episodes that reach the goal within the action budget. SPL weights each success by $L_{\rm ref}/\max(L_{\rm ref},L)$, where $L$ counts executed actions. Discrete references are shortest paths under the environment dynamics. C2D uses geometric distance divided by a nominal step size. PointMass and Reacher use feasible paths from a reference controller. Thus C2D SPL measures efficiency relative to nominal straight-line travel, and native SPL uses the controller path as its reference.

  \paragraph{Averaging and error bars.}
  We first average queries or episodes within each model and test target. For combined navigation curves, we average targets within each common training seed, then show the mean and sample standard deviation across seeds. KD first pools the recorded queries or episodes across rules within KD2 and KD4. Queries, episodes, and planning runs from one model all count toward the same seed.

  \section{Navigation and Goal Scoring}\label{app:control}
  Navigation uses a frozen WM at every context size. The planners differ in how they score its predicted states.

  \paragraph{Planning with the frozen WM.}
  The cross-entropy method (CEM) samples action sequences and predicts their image embeddings through the WM. It scores them against a goal image and refits the sampling distribution to the best sequences. It executes the first action and replans from the observed image. Scalar-CEM plans using the learned distance head. Its predictions and scores use public images, actions, and context records.

  Discrete CEM simulates four actions per sequence, samples 128 sequences, and updates the distribution four times using the best 16. For scaling, MG scores mean rollout distance plus terminal distance. KD scores terminal distance and samples the executed action from the final first-action distribution. PointMass and Reacher plan 8 and 12 actions ahead. Table~\ref{tab:navigation-budgets} gives episode limits.

  \begingroup\centering\small
  \captionsetup{hypcap=false}\captionof{table}{\textbf{Action limits for reaching the goal.} Limits count actions executed in the environment and are shared across context sizes, planners, and static or online evaluation within each study.}\label{tab:navigation-budgets}
  \begin{tabular}{@{}lr@{}}\toprule
    Study                                                    & Actions       \\\midrule
    MG scaling                                               & 60            \\
    MG familiar-rule control                                 & 28            \\
    MG unseen pairs                                          & 45            \\
    MG context interventions (Table~\ref{tab:interventions}) & 28            \\
    MG parameter transfer                                    & 30, 45, or 60 \\
    MG higher-order combinations                             & 30--90        \\
    KD ICL and scaling                                       & 256           \\
    C2D parameter transfer                                   & 20            \\
    C2D combinations                                         & 15--60        \\
    PointMass                                                & 60            \\
    Reacher                                                  & 10            \\\bottomrule
  \end{tabular}\par\endgroup

  \paragraph{Learned distances and oracle diagnostics.}
  For each scaling WM, we train one distance head and choose its checkpoint by the lowest val SmoothL1 loss. MG uses 2,304 train and 576 val layout--rule instances over nine single rules. KD uses 512 train and 128 val instances over twelve rules (Table~\ref{tab:training-test-settings}). The training label is the fewest actions from the true successor to the goal under its training rule, divided by 32. KD labels account for locks and held keys. We use the same head on held-out rules.

  MG ICL parameter, familiar-rule, and pair heads use shortest grid paths with ordinary one-cell moves. C2D uses geometric distances. PointMass and Reacher use distances in task coordinates. The head learns from these labels during training and reads predicted successor and goal embeddings at evaluation.

  Oracle-CEM retrieves the rendered state closest to each predicted embedding and scores it with precomputed goal distances. Oracle-greedy does this for each possible next action and picks the smallest distance, breaking ties uniformly at random. The oracle methods use evaluator state images and distances unavailable to Scalar-CEM. The same CEM settings can use different inference FLOPs for different WM sizes. Continuous tests use Scalar-CEM.

\end{AppendixProtocols}

\onecolumn
\AppendixPage{G1-MG-parameters}{
  \section{Results by Test Rule}\label{app:extended}
  \subsection{MG Parameter Transfer}
  \AppendixText{
    Each MG parameter is held out from training and tested on 64 new maps using three trained WMs. Figure~\ref{fig:mg-parameter-details} shows performance across context sizes. Table~\ref{tab:parameter-holdouts} reports endpoints and compares with retrieval using the same WMs. The field results include two WMs that fell short of an earlier val threshold.
  }
  \AppendixFigure[.985]{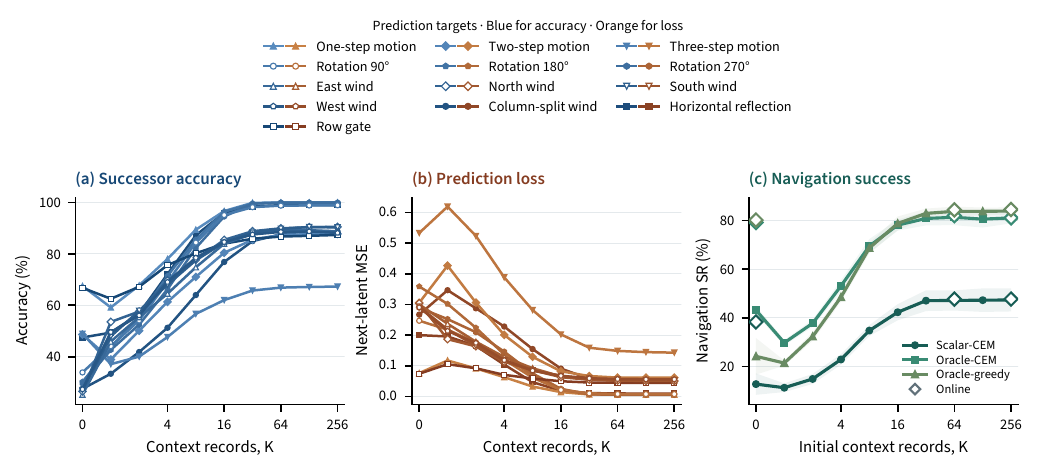}{%
    \textbf{MG results on 13 rule parameters withheld from training.} Panels (a) and (b) show test
    agent-cell accuracy and next-latent MSE, averaged over three WMs per target. Matching markers
    identify the same test rule. Panel~(c) shows goal-reaching success for the three planners,
    averaged over targets within each of three seeds. Bands are one sample standard deviation across
    seeds. Solid lines use fixed context. Hollow diamonds start at $K=0,64,256$ records and add
    observed transitions. Tasks allow 30, 45, or 60 actions.
  }{fig:mg-parameter-details}
  \par\begingroup\centering\footnotesize\setlength{\tabcolsep}{3pt}
  \captionsetup{hypcap=false}\captionof{table}{%
    \textbf{MG prediction and navigation by held-out parameter.} Entries average three WMs. Columns
    show test agent-cell accuracy and goal-reaching success at fixed $K=0$ and $K=256$ transition
    records. Retrieval at $K=256$ uses the WM encoder and a rule for picking a context transition
    tuned on val. The oracle planners use evaluator images and goal distances.
  }\label{tab:parameter-holdouts}
  \begin{tabular}{@{}lrrrrrrrrr@{}}
\toprule
\multirow{2}{*}{Target} & \multicolumn{2}{c}{WM accuracy} & Retrieval & \multicolumn{2}{c}{Scalar-CEM SR} & \multicolumn{2}{c}{Oracle-CEM SR} & \multicolumn{2}{c}{Oracle-greedy SR} \\
 & $K=0$ & $256$ & $256$ & $0$ & $256$ & $0$ & $256$ & $0$ & $256$ \\
\midrule
One-step motion & 67.5 & 100.0 & 71.4 & 18.8 & 63.5 & 97.4 & 99.5 & 50.5 & 100.0 \\
Two-step motion & 48.6 & 87.8 & 53.9 & 3.1 & 8.9 & 4.2 & 40.6 & 0.5 & 33.9 \\
Three-step motion & 48.9 & 67.3 & 51.8 & 1.6 & 4.2 & 8.9 & 10.4 & 3.6 & 3.1 \\
Rotation $90^\circ$ & 33.9 & 98.9 & 71.2 & 0.0 & 65.1 & 0.0 & 99.5 & 0.0 & 100.0 \\
Rotation $180^\circ$ & 30.2 & 99.9 & 70.2 & 0.0 & 69.3 & 0.0 & 100.0 & 0.0 & 100.0 \\
Rotation $270^\circ$ & 29.6 & 99.9 & 69.5 & 0.0 & 74.0 & 0.0 & 99.5 & 0.0 & 100.0 \\
North wind & 26.5 & 90.5 & 47.5 & 31.8 & 60.4 & 77.6 & 83.9 & 54.7 & 92.2 \\
South wind & 27.0 & 90.3 & 47.3 & 32.3 & 56.8 & 80.2 & 89.1 & 48.4 & 94.8 \\
West wind & 27.3 & 88.1 & 47.7 & 24.0 & 43.2 & 82.3 & 84.4 & 57.3 & 91.7 \\
East wind & 25.3 & 88.7 & 46.8 & 20.8 & 51.0 & 81.2 & 81.8 & 31.8 & 92.2 \\
Column-split wind & 27.5 & 88.7 & 46.6 & 20.3 & 38.5 & 56.8 & 83.9 & 42.7 & 93.8 \\
Horizontal reflection & 47.5 & 99.0 & 69.0 & 0.0 & 49.5 & 0.0 & 96.9 & 0.0 & 100.0 \\
Row gate & 66.8 & 87.3 & 72.3 & 13.0 & 30.7 & 71.9 & 84.4 & 24.5 & 88.5 \\
\bottomrule
\end{tabular}
\par\endgroup\medskip%

  \AppendixColumns{
    \paragraph{Predicting by retrieval.}
    Using the same image encoder and context, this baseline finds the closest context transition. It either copies that record's successor embedding or adds its embedding change to the query embedding. Val chooses the variant and action weight. With 256 records, the WM has higher accuracy on all 13 test parameters.
  }{
    \paragraph{Missed row-gate blocks.}
    For the same three WMs, row-gate accuracy rises from 66.8\% to 87.3\% over all queries as $K$ goes from 0 to 256. On the 507 test queries per WM where the row gate blocks horizontal motion, accuracy falls from 52.6\% to 0.4\%. This subset checks whether the model predicts the required stop.
  }
}

\AppendixPage{G2-MG-combinations}{
  \subsection{MG Transfer to Unseen Combinations}
  \AppendixText{
    For each seed, one WM trains on single rules and is tested on all ten unseen pairs (Fig.~\ref{fig:mg-pair-details}). Each three- to five-rule target uses its own WMs (Fig.~\ref{fig:mg-higher-order-details}). Training uses the constituent rules and their smaller combinations, with additional variants for some targets. The complete test combination remains held out.
  }
  \AppendixFigure{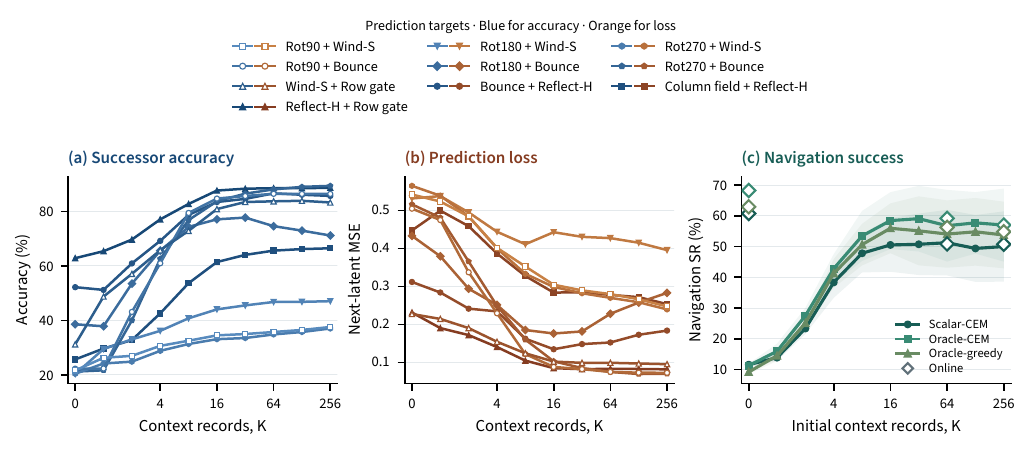}{%
    \textbf{MG results on ten unseen pairs of movement rules.} Test agent-cell accuracy (a) and
    next-latent MSE (b) use the same three WMs for all pairs. Panel~(c) shows the three planners. We
    average pairs within each seed, then plot the mean and one sample standard deviation across the
    three shared seeds. Solid lines fix the context and hollow diamonds start at $K=0,64,256$
    records and add observed transitions. Each test task allows 45 actions.
  }{fig:mg-pair-details}
  \AppendixFigure{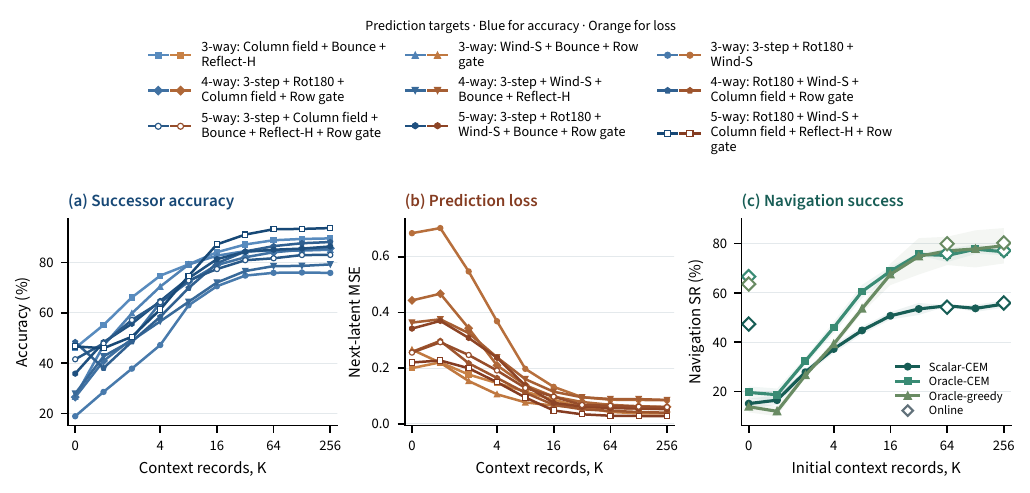}{%
    \textbf{MG results on nine unseen combinations of three to five movement rules.} Panels (a) and
    (b) show test agent-cell accuracy and next-latent MSE. Each test rule has three independently
    trained WMs. In (c), curves average navigation success across combinations within each seed, and
    bands show one sample standard deviation across seeds. Solid lines fix context. Hollow diamonds
    start at $K=0,64,256$ records and add observed transitions. Tasks allow 30--90 actions. Wind-S
    is south wind and Reflect-H swaps left and right.
  }{fig:mg-higher-order-details}
}

\AppendixPage{G3-KD}{
  \subsection{KeyDoor with One or Two Keys}
  \AppendixText{
    KD1 uses one key, trains on eleven altered rules, and tests ordinary one-cell motion. KD2 adds ordinary motion to train and tests the four combinations in Table~\ref{tab:training-test-settings}. KD3 and KD4 repeat those splits with two keys. From $K=0$ to 256, test accuracy of position and inventory rises from 29.4\% to 94.0\% (KD1), 23.2\% to 94.4\% (KD2), 31.5\% to 92.6\% (KD3), and 23.6\% to 93.2\% (KD4).
  }
  \AppendixFigure[.985]{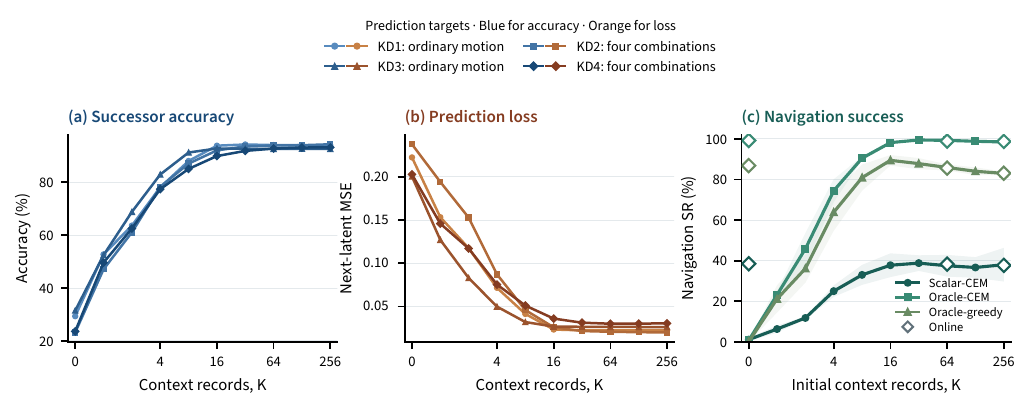}{%
    \textbf{KeyDoor results with one key (KD1, KD2) or two keys (KD3, KD4).} KD1 and KD3 test
    ordinary motion withheld from training. KD2 and KD4 test four unseen combinations. Panels (a)
    and (b) show joint cell-and-inventory accuracy and next-latent MSE over three models per study.
    For (c), we pool episodes within each study, average the four studies within each seed, and show
    one sample standard deviation across seeds. Hollow diamonds start with $K=0,64,256$ records and
    add observed transitions. Solid lines fix context. Tasks allow 256 actions.
  }{fig:kd-icl-details}
  \subsection{Continuous2D: Parameters and Combinations}
  \AppendixText{
    The two held-out values in each C2D parameter family (horizontal scale, drift, or right-half rotation) share three WMs. Each combined test rule uses its own three WMs.
  }
  \AppendixFigure[.985]{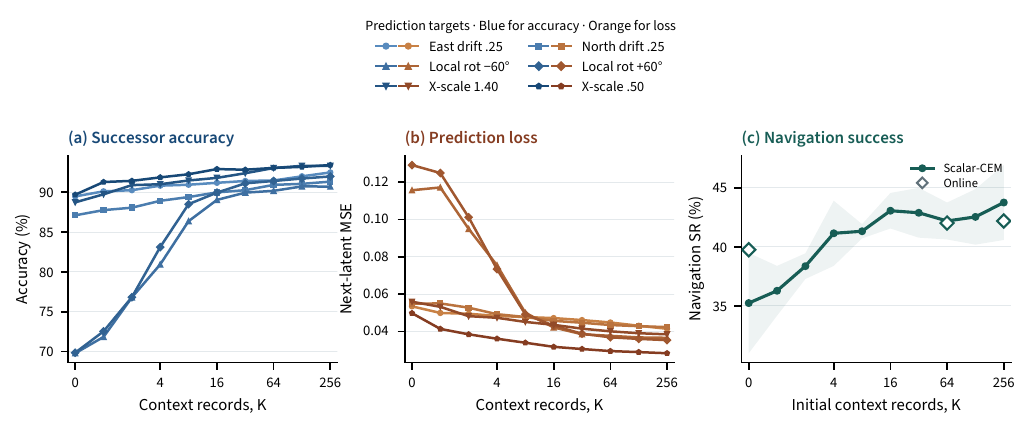}{%
    \textbf{Continuous2D results on six unseen parameter values.} Panels (a) and (b) show test
    position accuracy from retrieval among 64 successor images and next-latent MSE, averaged over
    three models per target. In (c), Scalar-CEM success averages targets within each seed, with one
    sample standard deviation across seeds. Solid lines fix context. Diamonds start at $K=0,64,256$
    records and add observed transitions. Tasks allow 20 actions.
  }{fig:c2d-parameter-icl}
}

\AppendixPage{G4-C2D}{
  \subsubsection*{C2D combinations}
  \AppendixFigure{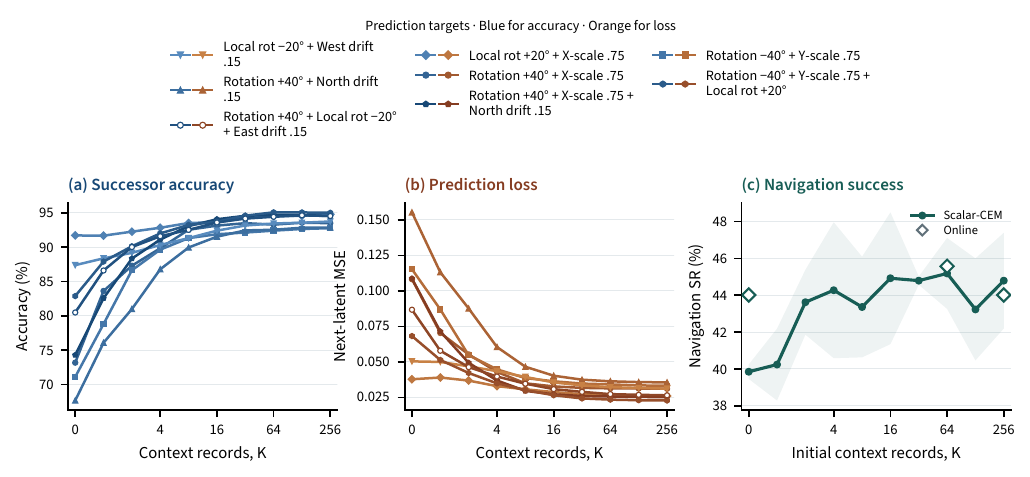}{%
    \textbf{Continuous2D results on five unseen pairs and three unseen triples.} Panels (a) and (b)
    show test position accuracy from retrieval among 64 successor images and next-latent MSE,
    averaged over three WMs per rule. In (c), Scalar-CEM success averages rules within each seed,
    with one sample standard deviation across seeds. Solid lines fix context. Diamonds start at
    $K=0,64,256$ records and add observed transitions. Tasks allow 15--60 actions.
  }{fig:c2d-composition-prediction}
  \subsection{PointMass and Reacher: Held-Out Force Gains}
  \AppendixText{
    Both simulators train on actuator gains $-3,-2,+2,+3$ and test gains $-4,+4$. Gain magnitude scales the force. A negative gain reverses its direction. Retrieval candidates share the current frame and differ in the future frame. Static and online navigation use the same weights and action limits. Reacher's ten-action budget failed the val check. Its test results are diagnostic.
  }
  \AppendixFigure{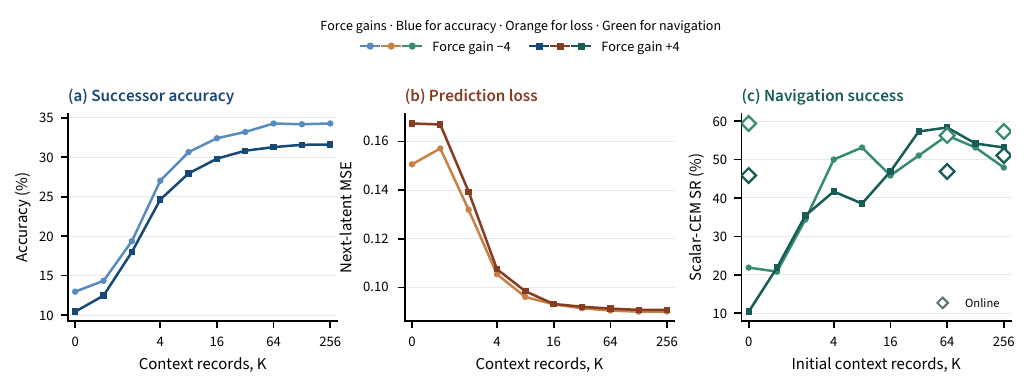}{%
    \textbf{PointMass results at held-out force gains $-4$ and $+4$.} Panels show future-frame
    retrieval accuracy among 64 candidates, next-latent MSE, and Scalar-CEM goal-reaching success.
    Matching markers link each gain across panels. Each curve averages the same three WMs for that
    gain. In (c), solid lines fix context and hollow diamonds start with $K=0,64,256$ records and
    add observed transitions. Tasks allow 60 actions.
  }{fig:pointmass-icl-details}
}

\AppendixPage{G5-native}{
  \AppendixFigure{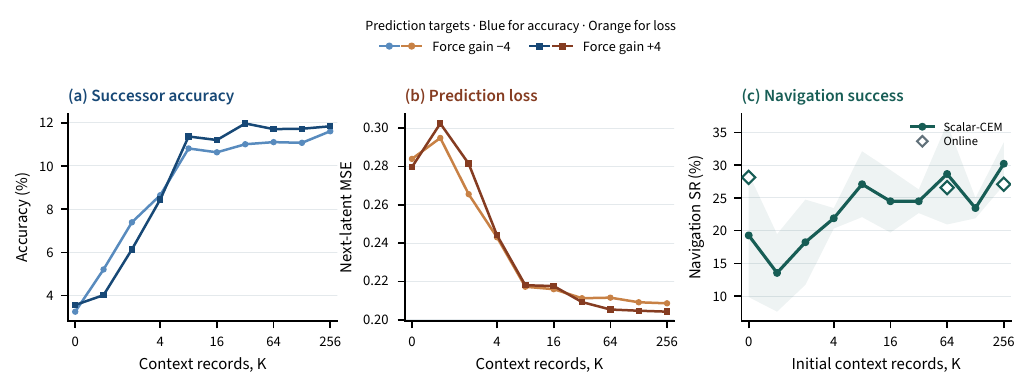}{%
    \textbf{Reacher results at held-out actuator gains $-4$ and $+4$.} Panels (a) and (b) show
    future-frame retrieval accuracy among 64 candidates and next-latent MSE, averaged over three
    WMs. Panel~(c) averages Scalar-CEM success over gains within each seed and shows one sample
    standard deviation across seeds. Solid lines fix context. Diamonds start with $K=0,64,256$
    records and add observed transitions. The ten-action limit did not pass the val budget check.
  }{fig:reacher-icl}
  \subsection*{Parameter transfer across continuous environments}
  \AppendixText{
    Table~\ref{tab:continuous} reports prediction and navigation at $K=0$ and $K=256$ for each held-out continuous parameter.
  }
  \par\begingroup\centering\footnotesize\setlength{\tabcolsep}{3pt}
  \captionsetup{hypcap=false}\captionof{table}{%
    \textbf{Continuous test results with no context or 256 fixed transition records.} Values are
    percentages (mean $\pm$ sample standard deviation across three WMs). Retrieval uses 64 successor
    candidates (Appendix~\ref{app:metrics}). Scalar-CEM success is the fraction of tasks reaching
    the goal within 20 actions for C2D, 60 for PointMass, or 10 for the Reacher diagnostic.
  }\label{tab:continuous}
  \begin{tabular}{@{}llrrrr@{}}
\toprule
\multirow{2}{*}{Environment} & \multirow{2}{*}{Held-out parameter} & \multicolumn{2}{c}{Successor retrieval (\%)} & \multicolumn{2}{c}{Scalar-CEM SR (\%)} \\
 & & $K=0$ & $256$ & $K=0$ & $256$ \\
\midrule
\multirow{6}{*}{\CtwoDabbr} & East drift $0.25$ & 89.5 $\pm$ 0.9 & 92.5 $\pm$ 0.7 & 45.8 $\pm$ 6.5 & 47.9 $\pm$ 4.8 \\
 & North drift $0.25$ & 87.1 $\pm$ 1.0 & 91.3 $\pm$ 0.9 & 44.8 $\pm$ 1.8 & 47.9 $\pm$ 1.8 \\
 & Local rotation $-60^\circ$ & 69.8 $\pm$ 2.8 & 90.7 $\pm$ 0.2 & 24.0 $\pm$ 6.5 & 41.7 $\pm$ 3.6 \\
 & Local rotation $+60^\circ$ & 69.8 $\pm$ 2.6 & 92.0 $\pm$ 0.2 & 17.7 $\pm$ 7.2 & 42.7 $\pm$ 3.6 \\
 & Horizontal scale $1.40$ & 88.7 $\pm$ 0.9 & 93.3 $\pm$ 0.8 & 38.5 $\pm$ 6.5 & 40.6 $\pm$ 8.3 \\
 & Horizontal scale $0.50$ & 89.7 $\pm$ 0.9 & 93.4 $\pm$ 0.3 & 40.6 $\pm$ 6.2 & 41.7 $\pm$ 3.6 \\
\midrule
\multirow{2}{*}{\PointMassname} & Force gain $-4$ & 13.0 $\pm$ 1.5 & 34.3 $\pm$ 4.7 & 21.9 $\pm$ 13.6 & 47.9 $\pm$ 6.5 \\
 & Force gain $+4$ & 10.4 $\pm$ 2.0 & 31.6 $\pm$ 5.6 & 10.4 $\pm$ 7.9 & 53.1 $\pm$ 6.2 \\
\midrule
\multirow{2}{*}{\Reachername} & Actuator gain $-4$ & 3.3 $\pm$ 0.2 & 11.6 $\pm$ 3.5 & 20.8 $\pm$ 11.0 & 28.1 $\pm$ 3.1 \\
 & Actuator gain $+4$ & 3.6 $\pm$ 0.9 & 11.8 $\pm$ 2.7 & 17.7 $\pm$ 14.4 & 32.3 $\pm$ 4.8 \\
\bottomrule
\end{tabular}
\par\endgroup\medskip%

  \section{Code Availability}\label{app:reproducibility}
  \noindent Code and reproduction instructions are available at \url{https://github.com/yuhaoyan04/scaling-to-adapt}.
}

\fi

\end{document}